\documentclass[runningheads]{llncs}

\usepackage{eccv}

\usepackage{eccvabbrv}

\usepackage{graphicx}
\usepackage{booktabs}

\usepackage[accsupp]{axessibility}  

\usepackage{hyperref}
\usepackage{adjustbox}
\usepackage{adjustbox}

\usepackage{orcidlink}

\newif\ifdraft
\draftfalse  
\ifdraft
    \newcommand{\syk}[1]{{\color{blue}[Soo Ye: #1]}}
    \newcommand{\sooye}[1]{\textcolor{blue}{#1}}
    \newcommand{\gct}[1]{{\color{purple}#1}}
\else
    \newcommand{\syk}[1]{}
    \newcommand{\sooye}[1]{\textcolor{black}{#1}}
    \newcommand{\gct}[1]{{\color{black}#1}}
\fi

\usepackage{pgfplotstable}
\definecolor{findOptimalPartition}{HTML}{FFCB7F}
\definecolor{storeClusterComponent}{HTML}{F5A42D}
\definecolor{dbscan}{HTML}{E0FF64}
\definecolor{constructCluster}{HTML}{B1DB0B}
\usepackage{multirow}
\usepackage{amssymb}
\usepackage{pifont}
\newcommand{\cmark}{\ding{51}}%
\newcommand{\xmark}{\ding{55}}%
\begin{document}

\title{Thinking Outside the BBox: \\Unconstrained Generative Object Compositing} 

\titlerunning{Thinking Outside the BBox}


\author{Gemma Canet Tarrés\inst{1}\orcidlink{0000-0001-5642-8282} \and Zhe Lin\inst{2}\orcidlink{0000-0003-1154-9907} \and Zhifei Zhang\inst{2}\orcidlink{0000-0003-0466-9548} \and Jianming Zhang\inst{2}\orcidlink{0000-0002-9954-6294} \and Yizhi Song\inst{3} \and Dan Ruta\inst{1}\orcidlink{0009-0000-0310-6933} \and Andrew Gilbert\inst{1}\orcidlink{0000-0003-3898-0596} \and John Collomosse\inst{1,2}\orcidlink{0000-0003-3580-4685} \and Soo Ye Kim\inst{2}\orcidlink{0009-0003-8104-4100}}

\authorrunning{G. Canet Tarrés et al.}

\institute{University of Surrey \email{\{g.canettarres,d.ruta,a.gilbert,j.collomosse\}@surrey.ac.uk} \and Adobe Research \email{\{zlin,zzhang,jianmzha,sooyek\}@adobe.com} \and Purdue University \email{song630@purdue.edu}}


\maketitle

\begin{figure}[h!]
    \centering
    \includegraphics[width=\linewidth]{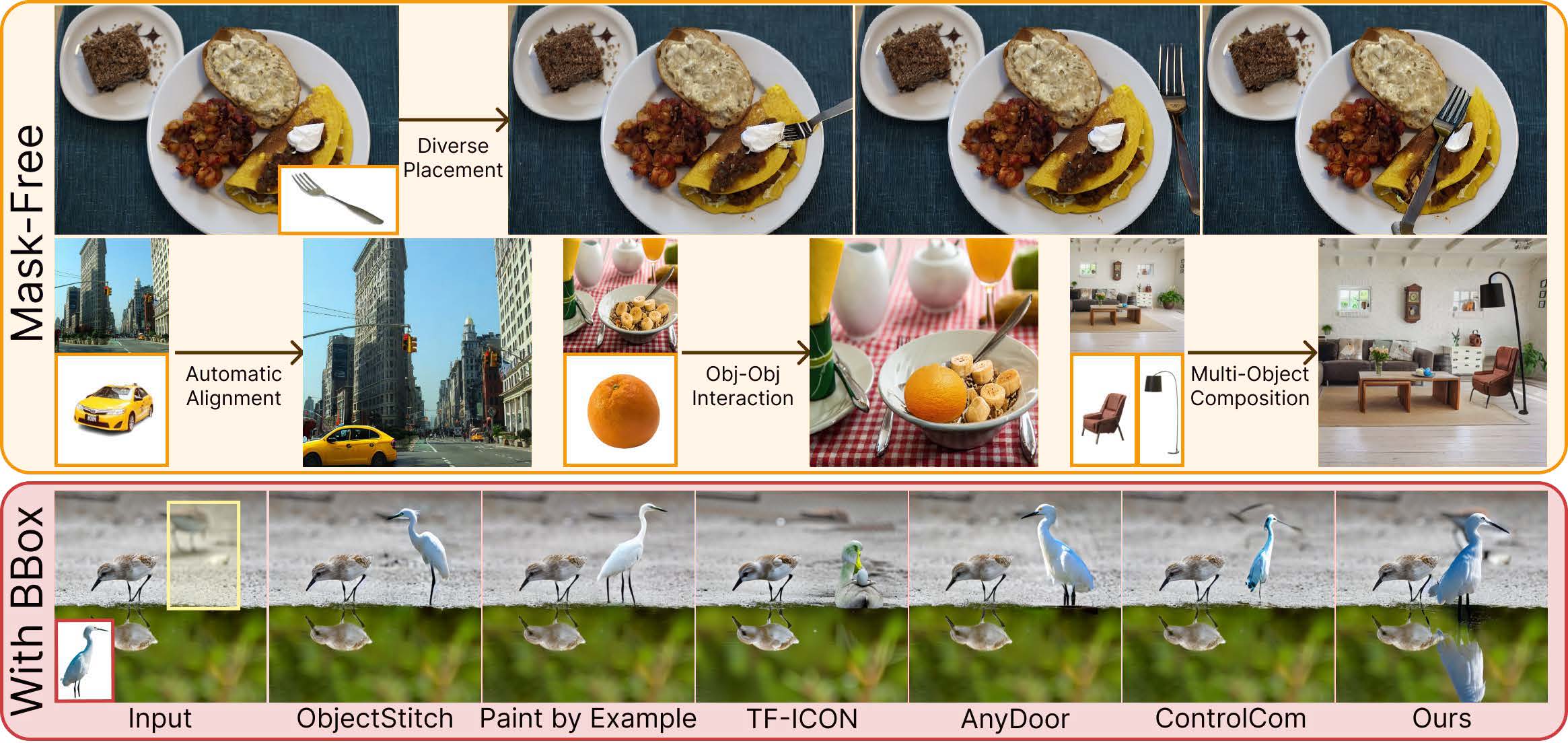}
    \caption{Our unconstrained object compositing model \sooye{has various advantages.}
    When using a bbox (\textit{bottom}), our model achieves \textbf{better background preservation} (see bird in background) and more \textbf{natural shadows and reflections} than SotA models \cite{song2022objectstitch,yang2023paintbyexample,lu2023tficon,chen2023anydoor,zhang2023controlcom} by allowing generation beyond the bbox. \sooye{Without \textbf{any} bbox input (\textit{top}), our model can automatically \textbf{place and composite} objects in diverse ways.}
    } 
    \label{fig:teaser}
\end{figure}

\begin{abstract}
Compositing an object into an image involves multiple non-trivial sub-tasks such as object placement and scaling, color/lighting harmonization, viewpoint/geometry adjustment, and shadow/reflection generation. Recent generative image compositing methods leverage diffusion models to handle multiple sub-tasks at once. However, existing models face limitations due to their reliance on masking the original object \sooye{during} training, which constrains their generation to the input mask. 
Furthermore, obtaining an accurate input mask specifying the location and scale of the object in a new image can be highly challenging. To overcome such limitations, we define a novel problem of \textit{unconstrained generative object compositing}, i.e., the generation is not bounded by the mask, and train a diffusion-based model on a synthesized paired dataset. Our first-of-its-kind model is able to generate object effects such as shadows and reflections that go beyond the mask, enhancing image realism. 
Additionally, if an empty mask is provided, our model automatically places the object in diverse natural locations and scales, accelerating the compositing workflow. Our model outperforms existing object placement and compositing models in various quality metrics and user studies.
\end{abstract}

\section{Introduction}
\label{sec:intro}


The rise of generative AI models and social media has sparked widespread interest in image editing techniques. Realistic and controllable image editing is now indispensable for applications like content creation, marketing, and entertainment. One pivotal step in most editing processes is image compositing, seamlessly integrating a foreground object with a background image. However, image compositing presents numerous challenges, including incorporating new shadows or reflections, illumination misalignment, unnatural foreground object boundaries, and ensuring the object's pose, location, and scale are semantically coherent. 

Previous works on image compositing \cite{liu2020arshadowgan,xu2017deep,xue2022dccf,lin2018st,azadi2020compositional} focus on specific sub-tasks, such as image blending, harmonization, object placement, or shadow generation. More recent methods \cite{song2022objectstitch,yang2023paintbyexample,chen2023anydoor, lu2023tficon} demonstrate that some separate compositing aspects (i.e., color harmonization, relighting, object geometry adjustments, and shadow/reflection generation) can be handled simultaneously using diffusion models \cite{rombach2022ldm,ho2020ddpm}. 
Such methods are often trained in a self-supervised way, masking an object in the ground truth image and using the masked image as input \cite{yang2023paintbyexample,chen2023anydoor} or denoising only within the mask region during the reverse diffusion process \cite{song2022objectstitch,chen2023anydoor}. Therefore, a mask is required as input during inference, leading to several limitations: (i) drawing an accurate mask can be non-trivial for an ordinary user and may result in unnatural composite images depending on the location, scale and shape of the input mask; (ii) the mask region constrains the generation, and their training data does not consider object effects, limiting the ability to synthesize appropriate effects such as long shadows and reflections; (iii) background areas near the object tend to be inconsistent with the original background, as the model does not see those regions if covered by the mask.

Therefore, \sooye{in this paper, we} propose a generative image compositing model that generates beyond the mask and even \textit{with an empty mask}, 
in which case the model automatically composites the object in a natural location at a suitable scale. Our model is the \textit{first} end-to-end solution for image compositing that simultaneously addresses all sub-tasks of image compositing, including object placement.
To achieve this, we first use image inpainting to create training data comprising image triplets (a foreground object, \sooye{a full} background image, and the ground truth composite image) while carefully considering shadows and reflections. 
Our model learns to produce creative, visually pleasing compositions suitable as recommendations or final composite results.

Our contributions are as follows:
\begin{itemize}
\item We introduce a novel task, \textit{unconstrained image compositing}, where the generation is not bounded by the input mask and can even occur without one. This allows the generation of realistic object effects (shadows and reflections) that go beyond the mask \sooye{while preserving the surrounding background}.
\item We design a first-of-its-kind diffusion model for unconstrained image compositing, employing multi-scale image embeddings to facilitate the generation of objects at various scales.
\item We propose a novel data generation pipeline for this task by synthesizing the full background image via image inpainting while considering reflections and shadows. This allows the 
mask to be empty,
in which case, our model is able to automatically place the object in a natural location and scale.
\item We evaluate our model thoroughly with various quality metrics and user studies and show that our model outperforms the state-of-the-art generative image compositing and object placement prediction models.
\end{itemize}
\section{Related Work}
\label{sec:relatedwork}
\textbf{Personalized Image Generation.} 
The landscape of image generation has transformed significantly with the introduction of diffusion models \cite{ho2020ddpm,rombach2022ldm}, which iteratively apply transformations to images, 
creating intricate, diverse, and realistic visual content. Novel approaches in personalized image generation, such as DreamBooth \cite{ruiz2023dreambooth}, explore finetuning vocabularies to define specific identities. \cite{chen2023disenbooth,gal2022imageisworth,lu2023dreamcom} follow the same idea, while \cite{shi2023instantbooth,jia2023taming,chen2023subjectdriven} leverage large-scale upstream training to eliminate the need for test-time finetuning. BLIP-Diffusion \cite{li2023blipdiffusion} proposes accelerating the process by aligning images with text.


\noindent
\textbf{Image Editing.} Diffusion models are also revolutionizing image editing. 
These models, paired with text prompts, enable methods like SDEdit \cite{sdedit}, which accepts various inputs (i.e. strokes, image patches) for local editing. Other techniques (InpaintAnything \cite{yu2023inpaint}, BlendedDiffusion \cite{blendeddiff}), utilize text to describe objects to be inserted into images. Semantically disentangling GAN latent spaces \cite{richardson2021encoding,bau2020semantic,alaluf2022hyperstyle} and employing semantic masks \cite{bau2020semantic,gu2019mask,ling2021editgan,wang2022pretraining} for localized edits have also paved the way for controllable editing. However, diffusion models extend these capabilities to more intricate and diverse editing tasks \cite{prompttoprompt,kim2022diffusionclip,liu2023more,imagic}.


\noindent
\textbf{Generative Image Compositing.} \gct{Early compositing methods depended on hand-crafted features \cite{lalonde2007photo} and rendering \cite{karsch2011rendering} or 3D techniques \cite{kholgade20143d}, often being tedious and limited. Nowadays, these methods are predominantly based on diffusion models.}
ObjectStitch \cite{song2022objectstitch} composites an object into an image based on the location and scale provided as a bounding box without extending the generation process outside of this bounded region. \cite{yang2023paintbyexample, kim2023paintbysketch} add more flexibility by enabling free-form mask inputs, which guide the generated object's location and shape. \gct{ControlCom \cite{zhang2023controlcom} offers controllability by independently performing blending, harmonization, view synthesis and \sooye{compositing} and CustomNet \cite{yuan2023customnet} allows control of viewpoint and location.}
AnyDoor \cite{chen2023anydoor} achieves improved identity preservation by separating local and global features and leveraging the benefits of DINO \cite{oquab2023dinov2} \gct{and \cite{seyfioglu2024diffusechoose} does it by using a secondary U-Net encoder}. \cite{kulal2023putting} focus on human generation to improve results. TF-ICON \cite{lu2023tficon} performs cross-domain compositing without additional training, finetuning, or optimization of pre-trained diffusion models. \gct{\cite{zhang2023phd} train a module that allows \sooye{compositing} with a frozen diffusion model.} Some methods, like DreamEdit \cite{li2023dreamedit}, draw inspiration from personalized image generation, employing iterative inpainting to composite foreground and background. However, the success of these methods is always upper-bounded by the accuracy of the input mask, and their ability to generate object elements such as shadows or reflections is also limited by it. By taking an \textit{unconstrained} compositing approach, our model can overcome these limitations. 


\noindent
\textbf{Object Placement.} Object placement is a crucial step in an automatic image compositing pipeline. 
Several methods try to predict semantically and visually suitable locations and sizes for compositing objects. Traditional approaches scan images, assuming depth-based scale or context similarity \cite{remez2018learning, fang2019instaboost, zhang2020and}. Deep learning methods like \cite{tan2018and,lee2018context,volokitin2020efficiently, dvornik2018modeling,dvornik2019importance} suggest placement only based on background image and object category, improving performance but with certain limitations. Instance-specific generative models enhance performance by predicting spatial transformations for the object via GANs \cite{tripathi2019terse} or transformers \cite{zhan2019spatial}. \cite{zhang2020placenet} promotes diversity by enforcing pairwise distances between predicted placements, while \cite{zhou2022graconet} does it by considering object placement as a graph completion task. SimOPA \cite{liu2021opa} introduces a discriminative approach, offering a score to validate composite images' rationality in foreground object placement. FOPA \cite{niu2022fopa} provides a more efficient method for this task at a specific scale. Alternatively, TopNet \cite{zhu2023topnet} deploys transformers for scale and location rationality predictions.

\section{Methodology}
\label{sec:method}

We define a novel problem of \textit{unconstrained generative object compositing}, where the generation is not restricted to the input mask when compositing an object into another image. The main reason why existing image compositing models focus on mask-bounded generation stems from their training data. These models often obtain the input background image by masking out the original object from the full image, similar to how image inpainting models gather their data. In other words, their problem can be regarded as a reference-object-based image inpainting problem.  However, employing diffusion models for denoising the entire image based on masked inputs does not inherently enable the synthesis of convincing object effects extending beyond the mask. Therefore, a new training data generation scheme is essential for obtaining a diffusion-based model addressing unconstrained generative object compositing.
Furthermore, upon observing the sensitivity of off-the-shelf image encoders to object scale, we integrate a multiscale embedding to help our model generate objects at various scales. We train the full framework in multiple stages to obtain optimal compositing performance.

\begin{figure}[t]
    \centering
    \includegraphics[width=0.85\linewidth,trim=0cm 2cm 1cm 0cm,clip]{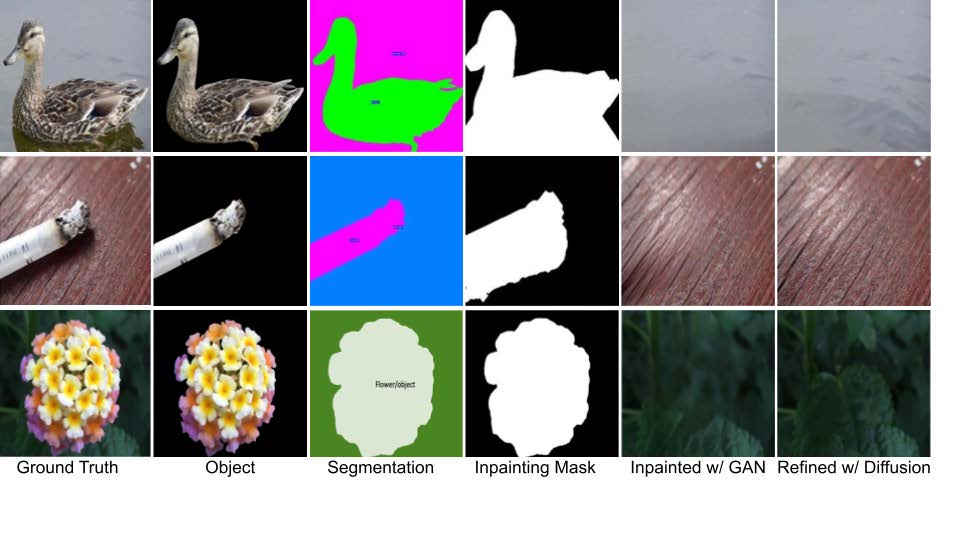}
    \caption{Visualization of the steps for synthesizing background images. Ground Truth corresponds to original images from \cite{pixabay}. \sooye{Our pipeline can be applied to any image.}}
    \label{fig:dataset}
\end{figure}

\begin{figure}[t]
    \centering
    \includegraphics[width=0.8\linewidth]{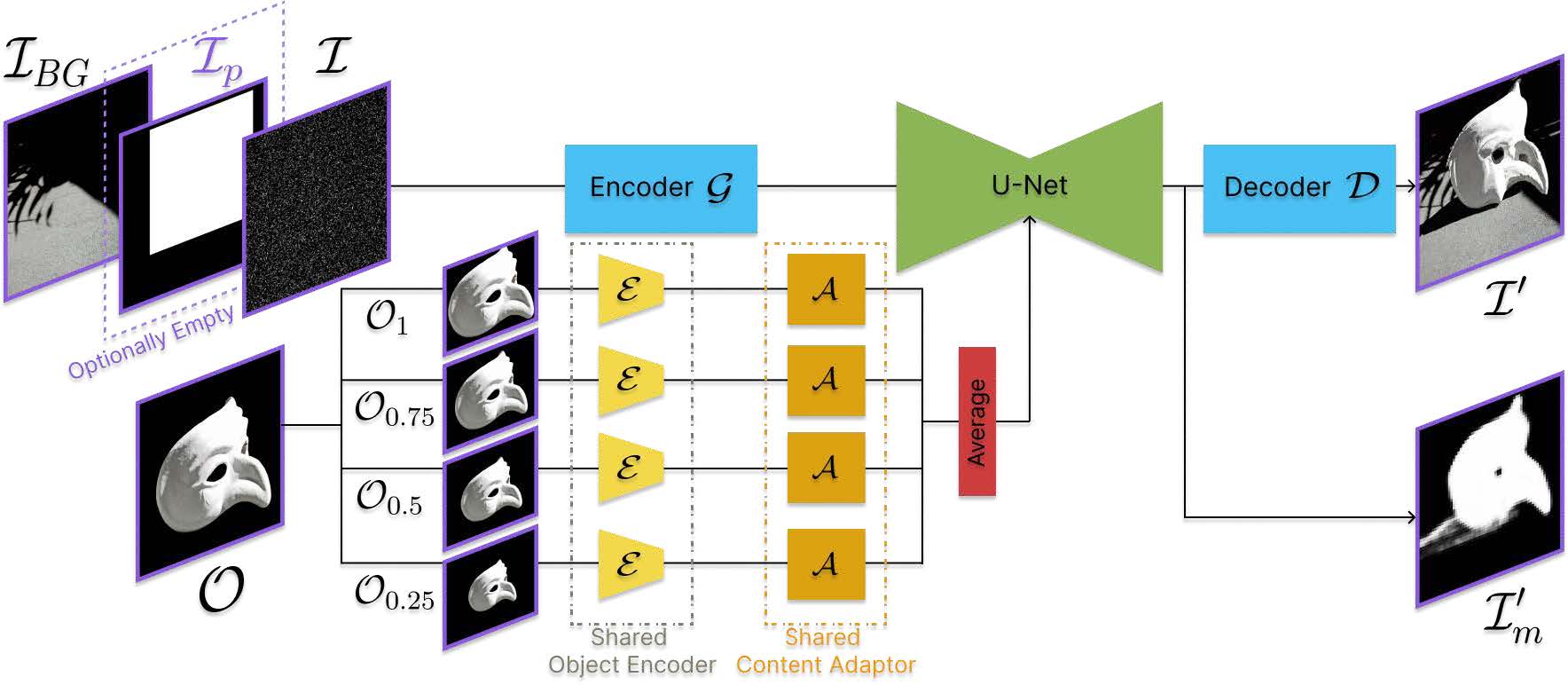}
    \caption{Model architecture. Our model consists of: (i) an object encoder $\mathcal{E}$ and a content adaptor $\mathcal{A}$ that encode the object at different scales; (ii) a Stable Diffusion backbone comprised of an autoencoder ($\mathcal{G}$, $\mathcal{D}$) and a U-Net. The multiscale \sooye{embeddings} from (i) are averaged to condition the U-Net via cross-attention. Background image $\mathcal{I}_{BG}$ and a mask $\mathcal{I}_{p}$ are concatenated to the input of (ii). $\mathcal{I}_{p}$ can be empty by setting all values to $-1$. \gct{The U-Net is adapted to return the predicted mask $\mathcal{I}'_m$ as an additional output.}}
    \label{fig:pipeline}
\end{figure}

\subsection{Data Generation Pipeline}
\label{sec:dataset}

Data generation is a crucial component of our method to enable unconstrained compositing. We avoid the necessity for masking the original image by going one step further and \textit{inpainting} the object within the original image. This approach yields a \textit{full} background image to be used as input to our model instead of a \textit{masked} background image, allowing us to eliminate the need for an input mask. 

Diffusion-based inpainting models \cite{rombach2022ldm,yang2020towards,avrahami2023blended,lugmayr2022repaint,xie2023smartbrush} 
have shown remarkable performance. However, they may occasionally introduce 
 \textit{generated new objects} inside the mask. 
To mitigate this, we first apply a GAN-based inpainting model \cite{zheng2022cmgan} to remove the object, then use a diffusion-based inpainting model on the GAN results with parameters set to preserve the underlying content. This sequential pipeline (Fig \ref{fig:dataset}) 
combines best of both worlds: consistent removal capability of GANs and enhanced inpainting quality of diffusion \sooye{(more details in SuppMat).}

Existing image compositing methods ignore object effects such as shadows and reflections when computing the input masked image, which limits 
their ability to synthesize such effects. For our model, we carefully consider shadows and reflections when calculating the inpainting mask for each image. 

Specifically, starting from any natural image, the following steps define the proposed data generation pipeline: \textbf{(i)} Segment the image via EntitySeg \cite{qi2022entityseg}, obtaining accurate foreground segmentation maps by filtering out those with low segmentation confidence score ($<30\%$). \textbf{(ii)} Filter out objects that are too large ($>80\%$ of the image) and have little background context or too small ($<1\%$) containing insignificant objects (e.g., small fragments of highly occluded objects). \textbf{(iii)} Apply an instance shadow detection model \cite{wang2022ssisv2} on a valid foreground object and retain shadows with confidence scores exceeding $80\%$ (middle row in Fig \ref{fig:dataset}). An exception is made for confidence scores exceeding $60\%$ if: (a) multiple objects of similar size are present, (b) they all have similar object-shadow vectors ($\sigma<2$), and (c) some of them have high confidence scores, in which case we consider the other shadows to have high confidence. \textbf{(iv)} For images involving a water body (filtered with segmentation label), the object mask is vertically flipped to obtain a reflection mask (top row in Fig \ref{fig:dataset}). \gct{More details \sooye{are provided} in SuppMat.} \textbf{(v)} The inpainting mask is defined as the union of object mask, shadow mask, and reflection mask, dilated by a $[40 \times 40]$ kernel to ensure every pixel is included. A GAN-based image inpainting model \cite{zheng2022cmgan} is applied, followed by a diffusion-based inpainting model, to synthesize the new background image.

For our training dataset, the above pipeline is applied for each of the 926K professional-level and license-free images collected from Pixabay \cite{pixabay} 
to obtain fully inpainted background images.

Another component of our training data sample is the object image. 
\sooye{We apply random transformations on the segmented object image in the same way as \cite{song2022objectstitch} to simulate real compositing scenarios.}
These include perspective warping and affine transformations to encourage varied pose and perspective of the object, and color shifting to learn correct relighting and color harmonization. 

\subsection{Model Architecture}
\label{sec:pipeline}
As shown in Fig \ref{fig:pipeline}, the background image $\mathcal{I}_{BG}$ and a mask $\mathcal{I}_{p}$ are concatenated with the 3-channel noise and fed into our model's backbone, Stable Diffusion 1.5 (SD)\cite{rombach2022ldm}, consisting of a variational autoencoder ($\mathcal{G}$, $\mathcal{D}$) and a U-Net. The foreground object $\mathcal{O}$ is processed by an object encoder $\mathcal{E}$ (CLIP ViT-L/14 \cite{radford2021clip}) and a content adaptor $\mathcal{A}$, aligning embeddings to text embeddings as in \cite{song2022objectstitch}.

\noindent
\textbf{Optional Position Mask \sooye{Input}.}
Our model can place objects in diverse natural locations and scales by feeding an empty mask (all values set to $-1$) instead of a position mask. This eliminates the laborious task of manual mask creation, accelerating the user workflow for image compositing. Moreover, the generated outputs can serve as location/scale suggestions to assist in creative ideation. 

However, having control over the generation can be useful if the user prefers a specific location and scale for the object. 
We train our model for enhanced controllability by alternating a bounding box position mask and an empty mask as the input mask.
Even when a bounding box is provided, it serves as a rough guide for the object's location and scale rather than a strict boundary as in existing compositing models. 
The generation is still \textit{unconstrained}, and the input background image is the \textit{full} inpainted image, not a masked one. 

Specifically, random perturbations are applied to the ground truth bounding box during training: the height and width are independently rescaled by a maximum of $\pm 10\%$, and a translation of $\leq 10$ pixels is applied to both axes. This results in a bounding box with perturbed location, scale, and aspect ratio. This strategy prevents the model from learning a rigid correspondence between the bounding box and the object's position, enabling adaptive adjustments. 

\noindent
\textbf{Mask Prediction for Enhanced Usability.}
The U-Net in our SD framework estimates an additional binary mask that indicates generated object pixels. This mask becomes usable in as few as 10 timesteps, much faster than the 50 timesteps needed for a full composite image. Though learned with the ground truth object mask, it also tends to include generated object effects. This mask can serve as an object placement prediction result and as a realistic input mask for other mask-based compositing models if desired. See SuppMat for more details.


\noindent
\label{sec:multiscale}
\textbf{Multiscale Object Encoding.} The core challenge of mask-free image compositing is accurately learning object locations and scales. This involves understanding inherent relative sizes and extrapolating apparent scale based on scene geometry. For example, a car may be larger than a person, but if placed farther in the background, it may appear smaller. Relying solely on training data diversity to learn these nuances is insufficient. We observe that $\mathcal{E}$ is scale-dependent (see Fig \ref{fig:multiscale}). Larger object encoders emphasize finer details, while smaller object encoders prioritize high-level structure. Thus, we incorporate multiscale encoding. The foreground object $\mathcal{O}$ undergoes bicubic downsampling by scale factors $s \in {1, 0.75, 0.5, 0.25}$. Each resized instance $\mathcal{O}_s$ is processed through the shared object encoder $\mathcal{E}$ and content adaptor $\mathcal{A}$, resulting in 4 embeddings. By averaging these, the model gains information of crucial object features at varying scales, enhancing its ability to determine the object's scale in the final image.

\subsection{Training Strategy}


\textbf{Multi-Stage Training.}
\label{sec:multistage}
Our model comprises different subnetworks: a U-Net, an image autoencoder ($\mathcal{G}$, $\mathcal{D}$), an object encoder $\mathcal{E}$, and a content adaptor $\mathcal{A}$. 
To train the network to understand accurate locations and scales without needing a mask, an empty mask with all values set to $-1$ replaces $\mathcal{I}_p$ in the initial training stages. 
Notably, several training stages are required to reach our final model.

1) The object encoder $\mathcal{E}$ is frozen while both the U-Net and content adaptor $\mathcal{A}$ are trained until convergence. This first stage model (\textit{S1}) generates highly diverse compositions but struggles with foreground object identity preservation.

2) To improve identity preservation, $\mathcal{E}$ is fine-tuned on multiview \cite{yu2023mvimgnet} and video data \cite{xu2018youtube, miao2022large}, using the paired data generation scheme from \cite{song2024imprint}. 
Then, $\mathcal{E}$ is frozen, and U-Net and $\mathcal{A}$ are trained until convergence (Model \textit{S2}). This stage enhances object identity preservation but diversity is compromised. 

3) Model merging \cite{wang2019merge} combines weights from \textit{S1} and \textit{S2} to achieve a balanced model (\textit{S3} = $0.25\cdot$\textit{S1} + $0.75\cdot$ \textit{S2}), optimizing diversity and identity preservation. The U-Net is then fine-tuned to ensure compatibility with the merged weights.



4) With $50\%$ probability, $\mathcal{I}_p $ alternates between a perturbed ground-truth mask 
and an empty mask filled with $-1$. It provides \textit{S4}, a model with optional control over object positioning. Only the U-Net is fine-tuned in this stage.


5) While previous stages were trained using $\mathcal{O}_{1}$ (encoding of $\mathcal{O}$ at scale $s=1$), this stage incorporates multiscale information from $\mathcal{O}$, with the U-Net being the sole subnetwork trained. \textit{S5} results in a model with improved scale accuracy.


6) Finally, the entire network is frozen except for an extra output layer in the U-Net, estimating the mask of the generated object and object effects in the resulting image. \gct{This step adds extra functionality without altering performance.}

\noindent
\textbf{Training Objectives.}
Based on the Stable Diffusion model \cite{rombach2022ldm}, we fine-tune our model with the following loss function:

\begin{equation}
    \mathcal{L}_{d} = \mathbb{E}_{\mathcal{I}_{c}, t, a_o, \epsilon \sim \mathcal{N} (0,1)} \left[ \left\| \epsilon - \epsilon_\theta\left(\mathcal{I}_{c}, t, a_o\right) \right\|_2^2 \right],
\label{eq:lossunet}
\end{equation}
where $a_o = \sum_{s}\mathcal{A}(\mathcal{E}(\mathcal{O}_{s}))/4$, $\mathcal{I}_{c}=[\mathcal{I}_{t}, \mathcal{I}_{p}, \mathcal{I}_{BG}]$ ($\mathcal{I}_{t}$: noisy version of the input image $\mathcal{I}$ at timestep $t$, $\mathcal{I}_{p}$: binary mask, $\mathcal{I}_{BG}$: background image, $[\cdot]$: concatenation operation across the channel dimension), $\epsilon_\theta$:denoising model being optimized. 


Dice loss is employed to estimate object mask $\mathcal{I}'_m$ as an extra network output:

\begin{equation}
    \mathcal{L}_{m} = 1 - \frac{2 \cdot |\mathcal{I}_{m} \cap \mathcal{I}'_{m} |}{|\mathcal{I}_m| + |\mathcal{I}'_m|},
\label{eq:lossdice}
\end{equation}

where $\mathcal{I}_m$ is the ground-truth segmentation mask for the object in the training data. $\mathcal{L}_{m}$ is integrated into the total loss by scaling it with $\lambda = 0.01$. 

\section{Experiments}
\label{sec:experiments}

\textbf{Training Details.}
During training, the learning rate is set to $10^{-4}$ for $\mathcal{E}$ and $4 \times 10^{-5}$ for the U-Net. All training stages utilize 8 A100 GPUs with the Adam optimizer and an effective batch size of 1024, using gradient accummulation.

\noindent
\textbf{Evaluation Dataset.}
We collect a real-world test set, \textit{DreamBooth}, of 113 object-background pairs by coupling object images from \cite{ruiz2023dreambooth} with plausible background images from Pixabay. 
Additionally, to evaluate compositing quality against ground truth, we generate \textit{Pixabay-Comp}, a synthetic dataset of 1K images where the input object image is obtained by randomly perturbing the object. 
For evaluating object placement prediction, we rely on \textit{OPA} \cite{liu2021opa}, a test set with 11K paired foreground and background samples designed for this task.

\noindent
\textbf{Evaluation Metrics.}
 To assess the identity and semantics preservation of the foreground object, CLIP-Score \cite{hessel2021clipscore}, DINO-Score \cite{oquab2023dinov2}, and DreamSim \cite{fu2023dreamsim} are computed by comparing the input object image and a cropped region of the generated image around the composited object, following \cite{chen2023anydoor,yang2023paintbyexample}. On our synthetic dataset, where the ground-truth composite images are available, we also compute FID to measure the quality of the entire generated image. Regarding object placement prediction, we use SimOPA \cite{liu2021opa} as a rationality score rating the location and scale of the object in the OPA dataset. On Pixabay-Comp, where only one accepted position is provided (the ground-truth), we compute the maximum IoU between the ground truth mask and five masks predicted by the model. Using this maximum value, we calculate its mean value and the percentage of images with IoU $> 0.5$, following \cite{zhu2023topnet}. \gct{These two scores are also provided on OPA for a more well-rounded comparison.} 
 For this task, measuring the diversity in estimating various natural locations and scales is essential, as otherwise, models may resort to predicting highly similar locations.
 Thus, pairwise LPIPS is used to evaluate the \textit{diversity} of our model's position predictions by averaging it among 5 different outputs of the same model. In addition to the quantitative metrics, we perform user studies to evaluate each task and combination. \gct{Additional details about the user studies are provided in SuppMat.}

\begin{figure}[t]
    \centering
    \includegraphics[width=\textwidth]{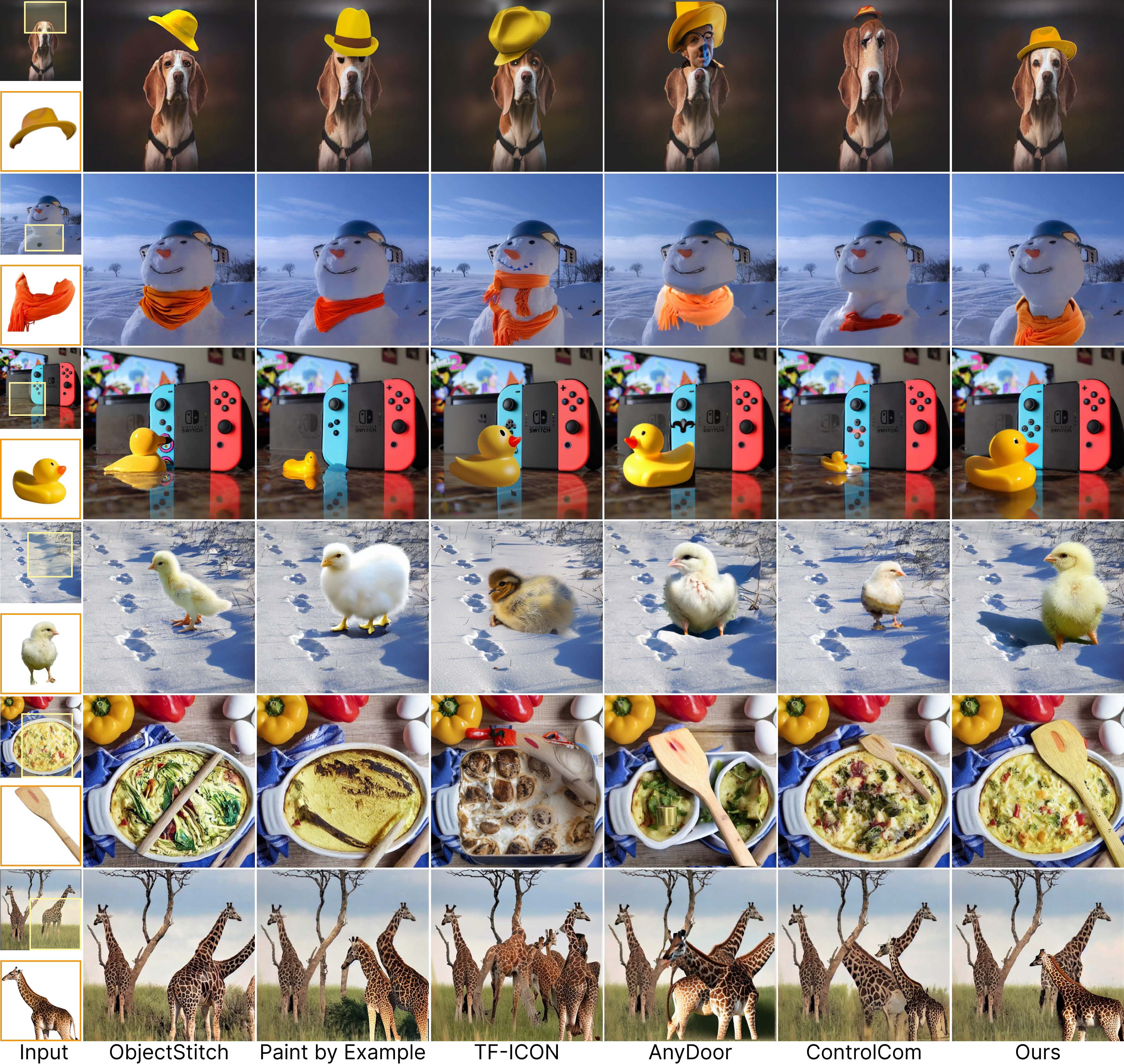}
    \caption{Visual comparison to generative image compositing models \cite{song2022objectstitch, yang2023paintbyexample, lu2023tficon, chen2023anydoor, zhang2023controlcom}, providing the same input object, background, and bbox. \sooye{Our model generates realistic results with natural shadows and reflections while preserving the original background.}} 
    \label{fig:baselines}
\end{figure}

\subsection{Comparison to Existing Methods}
\label{sec:baselines}
\gct{Our model can automatically place objects in natural locations while generating realistic composite images. Hence, we comprehensively evaluate our model for each task: Generative Object Compositing and Object Placement Prediction. We additionally compare to sequentially performing both tasks.}


\textbf{Generative Object Compositing.}
For this task, we compare to recent diffusion-based models: ObjectStitch \cite{song2022objectstitch}, Paint by Example \cite{yang2023paintbyexample}, TF-ICON \cite{lu2023tficon}, \gct{AnyDoor \cite{chen2023anydoor} and ControlCom \cite{zhang2023controlcom}}, which all require a mask input to guide the generation. Thus, we feed an input mask ($\mathcal{I}_{p}$) to our model as well.

\begin{figure}
    \centering
    \includegraphics[width=\textwidth]{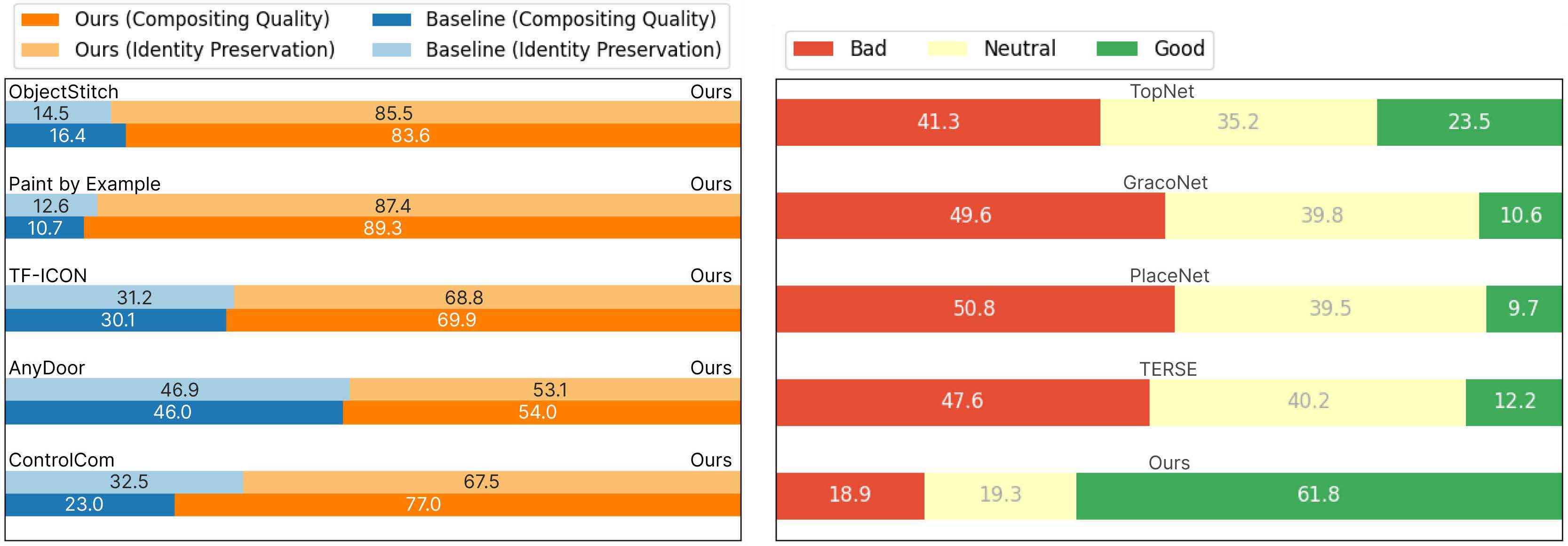}
    \caption{User studies. \textit{Left:} Percentage of users who chose our model or each baseline \cite{song2022objectstitch, yang2023paintbyexample, lu2023tficon, chen2023anydoor, zhang2023controlcom} in terms of compositing quality and identity preservation on DreamBooth. \textit{Right:} Users are asked to individually rate as Good, Bad or Neutral each location and scale prediction by Ours and compared models \cite{zhu2023topnet, zhang2020placenet, zhou2022graconet, tripathi2019terse} on a subset of OPA.}
    \label{fig:userstudies}
\end{figure}

All compared compositing models restrict generation to a part of the image defined by a mask. We zoom in on the generated area to boost their performance by expanding the bbox into a square with an amplifier ratio of 2.0, then resize it and paste it back to the original image, as in \cite{chen2023anydoor}. As shown in Tab \ref{tab:baselinesquality}, our model outperforms existing diffusion-based compositing methods \cite{song2022objectstitch, yang2023paintbyexample, lu2023tficon, chen2023anydoor, zhang2023controlcom} in
most quality metrics. To assess compositing quality and object identity preservation through subjective human evaluation, we performed a user study with 113 pairs of images in the DreamBooth test set shown to at least 5 users each, comparing our model's results with another model side by side, in randomized order. As shown in Fig \ref{fig:userstudies} (left), users preferred our model with up to $89.3\%$ preference rate in quality and $87.4\%$ in faithfulness of object identity.

\gct{Fig \ref{fig:baselines} showcases the different aspects in which our \textit{unconstrained model} is able to outperform mask-based state of the art image compositing models: (i) allowing the network to \sooye{leverage} information from the entire background image rather than just a masked area leads to \textbf{better background preservation} (rows 3 to 6); (ii) enabling object effects such as \textbf{shadows and reflections} beyond the bounding box allows for more natural and realistic composite images (rows 3-4); (iii) our model's success is not bounded by the accuracy of the bounding box thanks to its ability to adjust any \textbf{misaligned bounding box} (rows 1-2).}

\begin{table}[t!]
\centering
\begin{adjustbox}{width=\textwidth}

\begin{tabular}{lccccccc}
\toprule
\multirow{2}[3]{*}{\textbf{Method}} & \multicolumn{3}{c}{\textbf{DreamBooth}} & \multicolumn{4}{c}{\textbf{Pixabay-Comp}} \\
\cmidrule(lr){2-4} \cmidrule(lr){5-8}  & \textbf{CLIP-Score$\uparrow$} & \textbf{DINO-Score$\uparrow$} & \textbf{DreamSim$\downarrow$} & \textbf{FID$\downarrow$} & \textbf{CLIP-Score$\uparrow$} & \textbf{DINO-Score$\uparrow$} & \textbf{DreamSim$\downarrow$}\\ 
\cmidrule{1-8}
ObjectStitch$^\dag$ \cite{song2022objectstitch}                                                                                                           &      78.018                   &  85.247   &           0.342                  &      70.111                                                                                              &        74.964             &  77.506   &      0.488                \\
\cmidrule{1-8}
PaintByExample$^\dag$ \cite{yang2023paintbyexample}                                                                                                          &      77.782                   &  79.887   &           0.438                  &      82.923                                                                                              &        76.604             &  75.707   &      0.515                \\
\cmidrule{1-8}
TF-ICON$^\star$ \cite{lu2023tficon}                                                                                                           &      79.094                   &  81.781   &           0.341                  &      77.368                                                                                              &        75.694             &  77.810   &      0.485                \\
\cmidrule{1-8}
AnyDoor$^\ddag$ \cite{chen2023anydoor}                                                                                        &       80.619          & 83.632 &  \textbf{0.272}       &     72.996                                                                 &       \textbf{80.284}            &   80.829    &               0.399         \\ 
\cmidrule{1-8}
ControlCom$^\diamond$ \cite{zhang2023controlcom}                                                                                          &      74.312          &  70.497 &          0.424      &     66.071                                                                 &        72.006           &   67.476    &                    0.614   \\ 
\cmidrule{1-8}
Ours (w/ bbox)                                                                                           &    \textbf{80.946}      &  \textbf{85.646}   &       0.285   &   \textbf{62.406}      &                                                          77.129             &       \textbf{80.896}             &    \textbf{0.395}              \\ 

\bottomrule
\end{tabular}
\end{adjustbox}
\caption{Quantitative comparison of composition quality and identity preservation. FID is only computed on Pixabay-Comp, which has ground truth images. \gct{$^\dag$: Model finetuned on the same data as Ours. $^\ddag$: Paper version, already includes diverse video and multiview data. $^\star$: Paper version, inference-based model \sooye{that does not require training.} $^\diamond$: Paper version, no available \sooye{training} code.}} 
\label{tab:baselinesquality}
\end{table}

\begin{figure}[t]
    \centering
    \includegraphics[width=\linewidth]{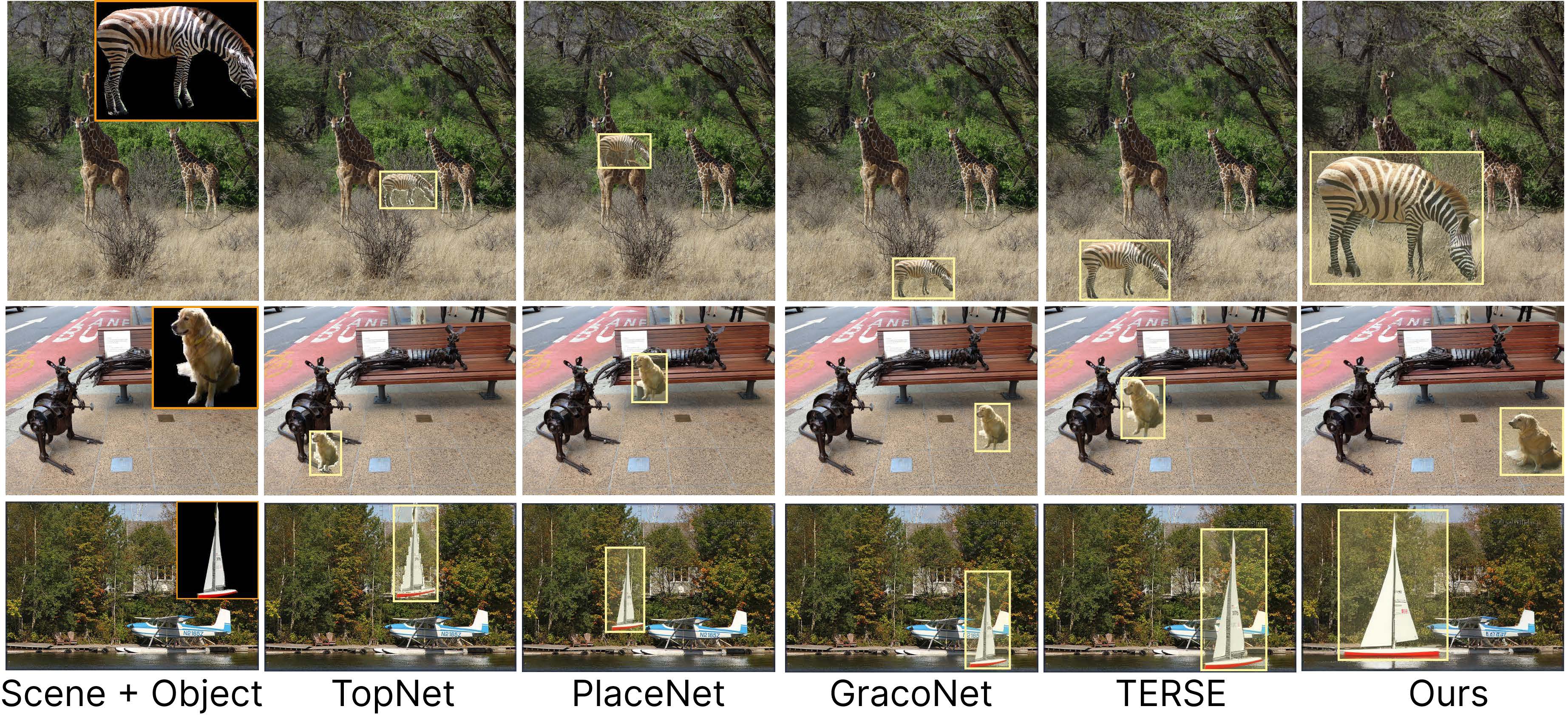}
    \caption{Visualization of location and scale prediction (marked in yellow) of our model and prior object placement prediction works \cite{zhu2023topnet, zhang2020placenet, zhou2022graconet, tripathi2019terse}. For visualization purpose, we display our generated image for our model and copy-paste the object into the predicted bounding box for \sooye{the compared models}.}
    \label{fig:baselinespos}
\end{figure}

\begin{table}[t!]
\centering
\begin{adjustbox}{width=\textwidth}
\begin{tabular}{lccccccc}
\toprule
\multirow{2}[3]{*}{\textbf{Method}} &  \multicolumn{4}{c}{\textbf{OPA}} & \multicolumn{3}{c}{\textbf{Pixabay-Comp}} \\
\cmidrule(lr){2-5} \cmidrule(lr){6-8} 

  & \textbf{SimOPA $\uparrow$} &     \textbf{LPIPS$\uparrow$}     &\textbf{IoU $>$ 0.5$\uparrow$} &  \textbf{mean-IoU$\uparrow$}          &\textbf{IoU $>$ 0.5$\uparrow$} &  \textbf{mean-IoU$\uparrow$} & \textbf{LPIPS$\uparrow$} \\ 
\cmidrule{1-8}
TopNet \cite{zhu2023topnet} &                                                                                                             0.256  &    2.758 & 16.8 \%  & 0.094          & 48.0 \% & 0.246 &  1.218  \\ 
\cmidrule{1-8}
GracoNet \cite{zhou2022graconet} &                                                                                                        \textbf{0.395}          &  0.836 & 12.2 \% & 0.189 & 30.2 \% & 0.327 &  2.832         \\ 
\cmidrule{1-8}
PlaceNet \cite{zhang2020placenet} &                                                                                                        0.197        &  0.746  & 11.2 \% & 0.194 & 8.6 \% & 0.237 & 2.072           \\ 
\cmidrule{1-8}
TERSE \cite{tripathi2019terse} &                                                                                          0.319                   &  0.000   & 10.8 \% & 0.123 & 12.2 \% &  0.230 &  0.000          \\ 
\cmidrule{1-8}
Ours (w/o bbox)          &  0.382   &   \textbf{5.619}      & \textbf{31.4 \%} & \textbf{0.196} &                                                       \textbf{65.4 \%}  & \textbf{0.562} & \textbf{3.158}                                           \\
\bottomrule
\end{tabular}
\end{adjustbox}

\caption{Quantitative evaluation of predicted location and scale of our model compared to state-of-the-art object placement prediction models. LPIPS is $ \times 10^{-3}$.}
\label{tab:baselineslocscale}
\end{table}

\noindent
\textbf{Object Placement Prediction.}
We compare our model to state-of-the-art object placement prediction models: TopNet \cite{zhu2023topnet}, GracoNet \cite{zhou2022graconet}, PlaceNet \cite{zhang2020placenet} and TERSE \cite{tripathi2019terse}, for evaluating object placement accuracy. Tab \ref{tab:baselineslocscale} shows our model's superior performance on all metrics for Pixabay-Comp and most metrics for the OPA dataset, with SimOPA results comparable to GracoNet \cite{zhou2022graconet}. Note that, unlike all other methods, our model was not trained on the OPA dataset as our final goal is object compositing. Additionally, we conducted a user study with 150 object-background pairs, simultaneously showing compositions from all five methods to at least 5 users, who rated each individual image as Good, Bad, or Neutral based on \textit{only} location and scale. Fig \ref{fig:userstudies} (right) shows our model's higher success rate. Visual results are in Fig \ref{fig:baselinespos}.

\begin{figure}[t]
    \centering
    \includegraphics[width=0.9\linewidth]{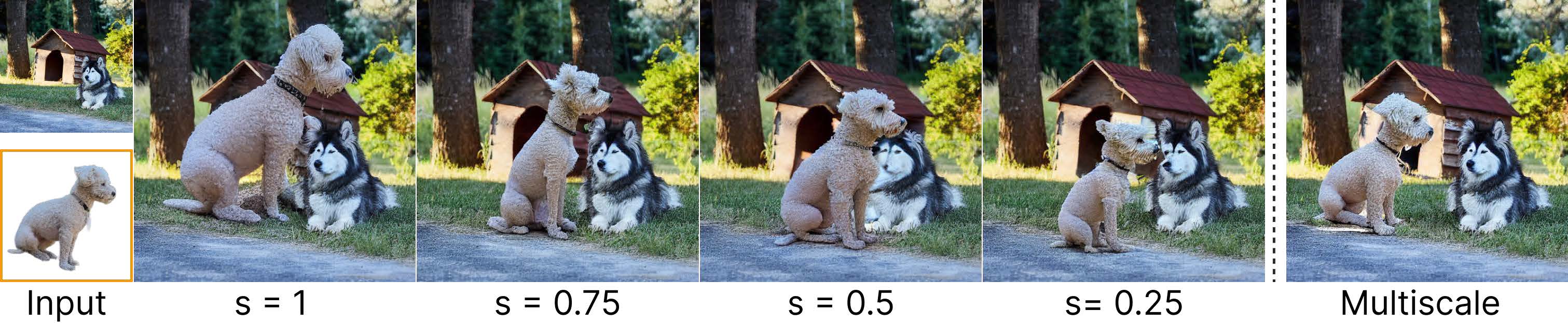}
    \caption{Visualization of scale effect on $\mathcal{E}$. The encoder is sensitive to object scale, making it challenging to generate smaller objects using just $s=1$. Our multiscale approach enhances object information, enabling diverse object scale generation.}
    \label{fig:multiscale}
\end{figure}

\noindent
\textbf{Automated Placement and Compositing Pipeline.}
\sooye{As our model is able to perform object placement and compositing in a joint manner, we further compare to a two-step pipeline by sequentially connecting TopNet \cite{zhu2023topnet} \gct{(best rated SotA model in user study)} with ObjectStitch \cite{song2022objectstitch}, Paint by Example \cite{yang2023paintbyexample}, TF-ICON \cite{lu2023tficon}, AnyDoor \cite{chen2023anydoor} and ControlCom \cite{zhang2023controlcom}. As shown in Table  \ref{tab:2stepsquant}, our model significantly outperforms each of the evaluated combinations on all quality metrics as well as a user study. The sequential approach is bound to error accumulation and can easily lead to unnatural results if the mask predictions are not precise. Please refer to SuppMat for visual examples.} 


\begin{table}[t!]

\centering
\begin{adjustbox}{width=\textwidth}

\begin{tabular}{lcccc}
\toprule
\textbf{Method}                           &  \textbf{CLIP-score$\uparrow$} & \textbf{DINO-score$\uparrow$} & \textbf{DreamSim$\downarrow$}  & \textbf{User Preference Over Ours}\\ 
\cmidrule{1-5}
TopNet \cite{zhu2023topnet} + ObjectStitch \cite{song2022objectstitch} &          54.900         & 58.110     & 0.756 &    19.20\%                                                                                                                                    \\ 
\cmidrule{1-5}
TopNet \cite{zhu2023topnet} + PaintByExample \cite{yang2023paintbyexample}          &  54.176   &   58.053  &  0.754 & 16.46\%                                                                                                       \\ 
\cmidrule{1-5}
TopNet \cite{zhu2023topnet} + TF-ICON \cite{lu2023tficon}          &  57.212   & 58.667  & 0.740  & 23.86\%                                                                              \\
\cmidrule{1-5}
TopNet \cite{zhu2023topnet} + AnyDoor \cite{chen2023anydoor}          &  56.040   &  59.305  & 0.718 &   21.06\%                                                                           \\ 
\cmidrule{1-5}
TopNet \cite{zhu2023topnet} + ControlCom \cite{zhang2023controlcom}          &   53.111  &  57.238 & 0.788 &  32.92\%                                                                                           \\ 
\cmidrule{1-5}
Ours (w/o bbox)          &   \textbf{69.597}  &  \textbf{71.527}  & \textbf{0.420} &  -                                                                                           \\ 

\bottomrule
\end{tabular}
\end{adjustbox}

\caption{Quantitative evaluation on quality and identity preservation of our mask-free model compared to a two-step pipeline combining TopNet \cite{zhu2023topnet} with SotA generative compositing models \cite{song2022objectstitch,yang2023paintbyexample,lu2023tficon,chen2023anydoor,zhang2023controlcom}. All metrics are computed on DreamBooth set.}
\label{tab:2stepsquant}
\end{table}

\begin{table}[t]

\centering
\begin{adjustbox}{width=\textwidth}

\begin{tabular}{lcccccccc}
\toprule

\textbf{} & \textbf{ Multiview/video} & \textbf{ Merged} & \textbf{ Optional bbox} & \textbf{ Multiscale}                           & \textbf{ FID$\downarrow$} & \textbf{ CLIP-score$\uparrow$} & \textbf{ DINO-score$\uparrow$} & \textbf{ DreamSim$\downarrow$} \\ 
\cmidrule{1-9}
\textit{(S1)} & \xmark & \xmark & \xmark & \xmark &      66.169             &   66.166   & 73.385 &    0.569                                                                                                                                            \\ 
\cmidrule{1-9}
\textit{(S2)}  & \cmark & \xmark & \xmark & \xmark          &   63.048  &   \textbf{67.177}  & \textbf{74.445}  & 0.529                                                                                                       \\ 
\cmidrule{1-9}
\textit{(S3)} & \cmark & \cmark & \xmark & \xmark           &  63.375   & 66.877  &  73.991 & 0.539                                                                              \\
\cmidrule{1-9}
\textit{(S4)} & \cmark & \cmark & \cmark & \xmark          &  62.918   &   66.806 & 74.059 &   0.538                                                                           \\ 
\cmidrule{1-9}
\textit{(S5)} & \cmark & \cmark & \cmark & \cmark          &   \textbf{62.406}  &  66.938  & 74.261 &  \textbf{0.528}                                                                                           \\ 

\bottomrule
\end{tabular}
\end{adjustbox}

\caption{Quantitative evaluation on quality and identity preservation of our model in each training stage, as defined in Sec \ref{sec:multistage}.}
\label{tab:ablationinitialcheckpoint}
\end{table}

\begin{figure}[t]
    \centering
    \includegraphics[width=\linewidth]{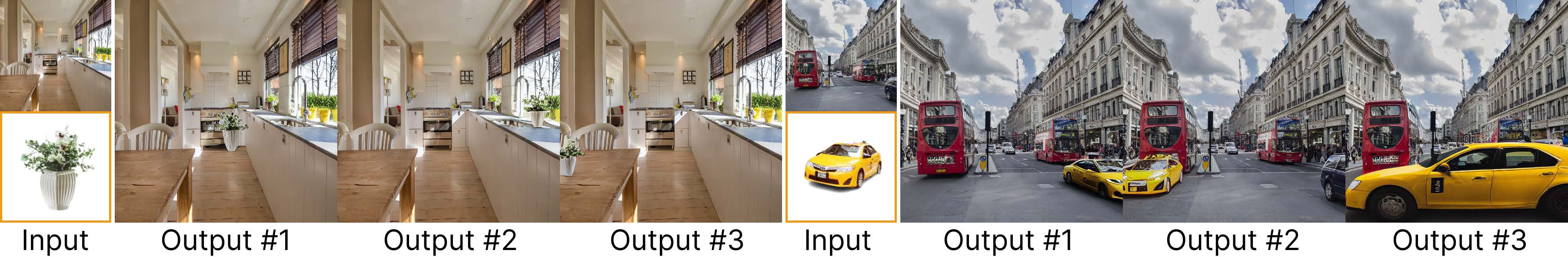}
    \caption{Visualizations of diverse compositions using our model with empty input mask. 
    }
    \label{fig:diversity}
\end{figure}





\begin{figure}[t!]
    \centering
    \includegraphics[width=\linewidth]{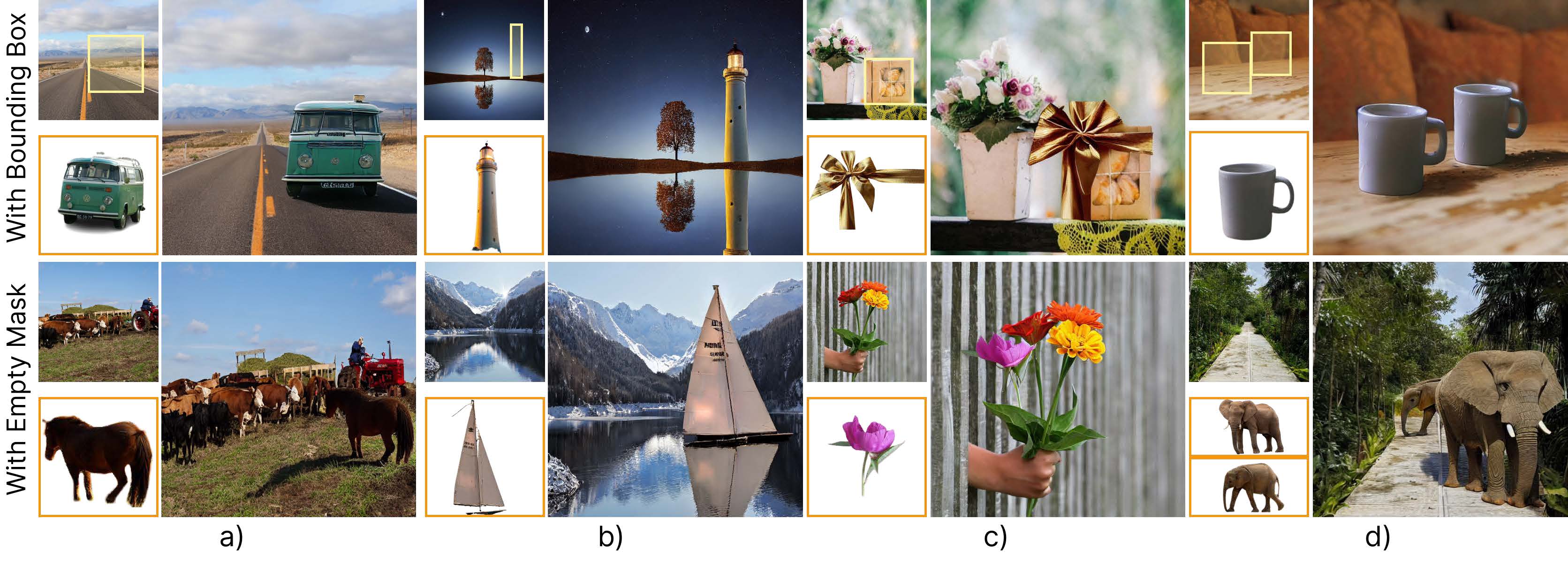}
    \caption{Examples of input and output for different applications. Top: a bbox mask is used to guide the location. Bottom: mask is empty and the model automatically places the object \sooye{at a natural location and scale} in the scene.
    }
    \label{fig:applications}
\end{figure}

\begin{figure}[t!]
    \centering
    \includegraphics[width=\linewidth]{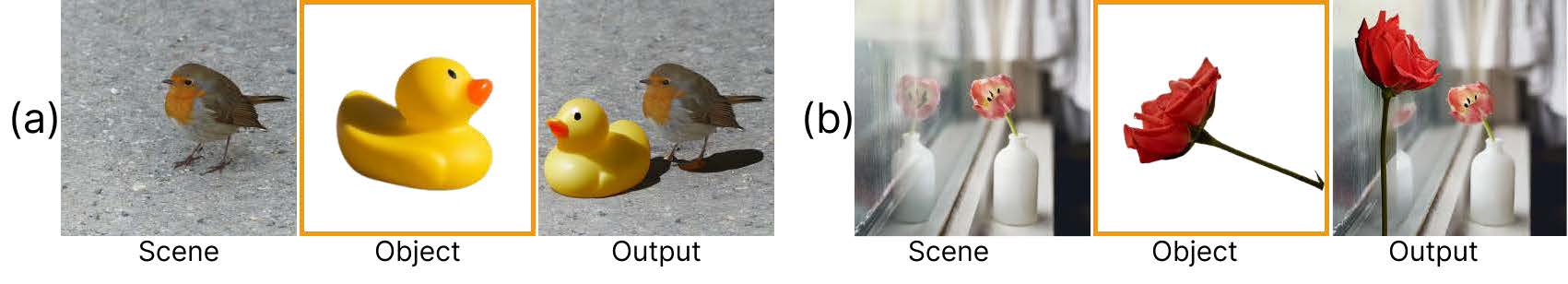}
    \caption{Visualization of two limitations of our model. (a): model harmonizes the \sooye{object in the} background image \sooye{along with} the input object in case of strong lighting on the input object. (b): lateral glass reflections are unaddressed. 
    }
    \label{fig:limitations}
\end{figure}

\subsection{Effect of Each Training Stage}


We analyze each training stage's impact on our model in Tab \ref{tab:ablationinitialcheckpoint}. Starting with the initial model (\textit{S1}) trained in our unconstrained compositing framework, each subsequent stage detailed in Sec \ref{sec:multistage} is incrementally integrated. Fine-tuning with multi-view and video data (\textit{S2}) improves identity preservation but limits object pose diversity, yielding unnatural results. Merging \textit{S2} with \textit{S1} in \textit{S3} retains benefits while broadening object pose variety. Introducing a rough bounding box input with $50\%$ probability in \textit{S4} allows the model to offer positioning guidance while also learning more precise object locations and scales. Our multiscale approach in \textit{S5} equips the model with more informative encodings of scale, resulting in semantically realistic compositions (Fig \ref{fig:multiscale}). More details in SuppMat.

\subsection{Applications}

\textbf{Creative Composite Image Recommendation.}
Our model can provide a diverse range of composite images varying the object's location, scale and pose (\sooye{Fig \ref{fig:teaser},} Fig \ref{fig:diversity}). 
Such results can serve as creative recommendations for the user.

\noindent
\textbf{Effortless Mask-Based Object Compositing.}
The unconstrained generation feature of our model allows for users to easily provide a rough desired location.
Even if this mask is not accurate, our model can naturally align the object and automatically add any necessary shadows (Fig \ref{fig:applications} a) and reflections (\sooye{Fig \ref{fig:teaser},} Fig \ref{fig:applications} b) \sooye{that extend beyond the mask}, resulting in a realistic composite image.




\noindent
\textbf{Object-Object Interaction.}
Our model's ability to generate beyond the bounding box enables easy addition of interacting objects in a scene without altering the surrounding background or existing objects 
(\sooye{ Fig \ref{fig:baselines} rows 1-2,} \gct{Fig \ref{fig:teaser}, }Fig \ref{fig:applications} c).


\noindent
\textbf{Multiple-Object Compositing.}
Our model is able to sequentially place different objects into a scene, ensuring that their relative location, scale and lighting are semantically and visually coherent (\sooye{Fig \ref{fig:teaser},} Fig \ref{fig:applications} d). 

\subsection{Limitations}
Fig \ref{fig:limitations} illustrates limitations of our model that we leave as future work. Compositing objects with hard lighting can sometimes cause unintended relighting of the entire scene rather than just the targeted object. Additionally, generating lateral reflections on glass remains challenging. Potential solutions include: (i) applying lighting perturbations on the object during training, and (ii) adding a symmetric object mask relative to the mirror plane during object inpainting.  

\section{Conclusion}
\label{sec:conclusions}
In this paper, we define a novel task of \textit{unconstrained object compositing} and propose \gct{the first} diffusion-based framework to address this task. Our model is carefully trained with a multi-stage strategy on synthetic paired data generated via inpainting while considering object effects. Through extensive evaluation across diverse datasets and metrics, we demonstrate state-of-the-art results in both 
object placement prediction and generative object compositing. Additionally, we show the benefits of unconstrained object compositing on various applications. We hope our paper can spur further research in this exciting direction.

\bibliographystyle{splncs04}
\bibliography{main}
\clearpage
\setcounter{page}{1}
\setcounter{section}{0}

\title{Thinking Outside the BBox: \\Unconstrained Generative Object Compositing} 
\
\titlerunning{Thinking Outside the BBox}


\author{Gemma Canet Tarrés\inst{1}\orcidlink{0000-0001-5642-8282} \and Zhe Lin\inst{2}\orcidlink{0000-0003-1154-9907} \and Zhifei Zhang\inst{2}\orcidlink{0000-0003-0466-9548} \and Jianming Zhang\inst{2}\orcidlink{0000-0002-9954-6294} \and Yizhi Song\inst{3} \and Dan Ruta\inst{1}\orcidlink{0009-0000-0310-6933} \and Andrew Gilbert\inst{1}\orcidlink{0000-0003-3898-0596} \and John Collomosse\inst{1,2}\orcidlink{0000-0003-3580-4685} \and Soo Ye Kim\inst{2}\orcidlink{0009-0003-8104-4100}}

\authorrunning{G. Canet Tarrés et al.}

\institute{University of Surrey \email{\{g.canettarres,d.ruta,a.gilbert,j.collomosse\}@surrey.ac.uk} \and Adobe Research \email{\{zlin,zzhang,jianmzha,sooyek\}@adobe.com} \and Purdue University \email{song630@purdue.edu}}

\maketitle
\begin{figure*}[t]
    \centering
    \includegraphics[width=\linewidth]{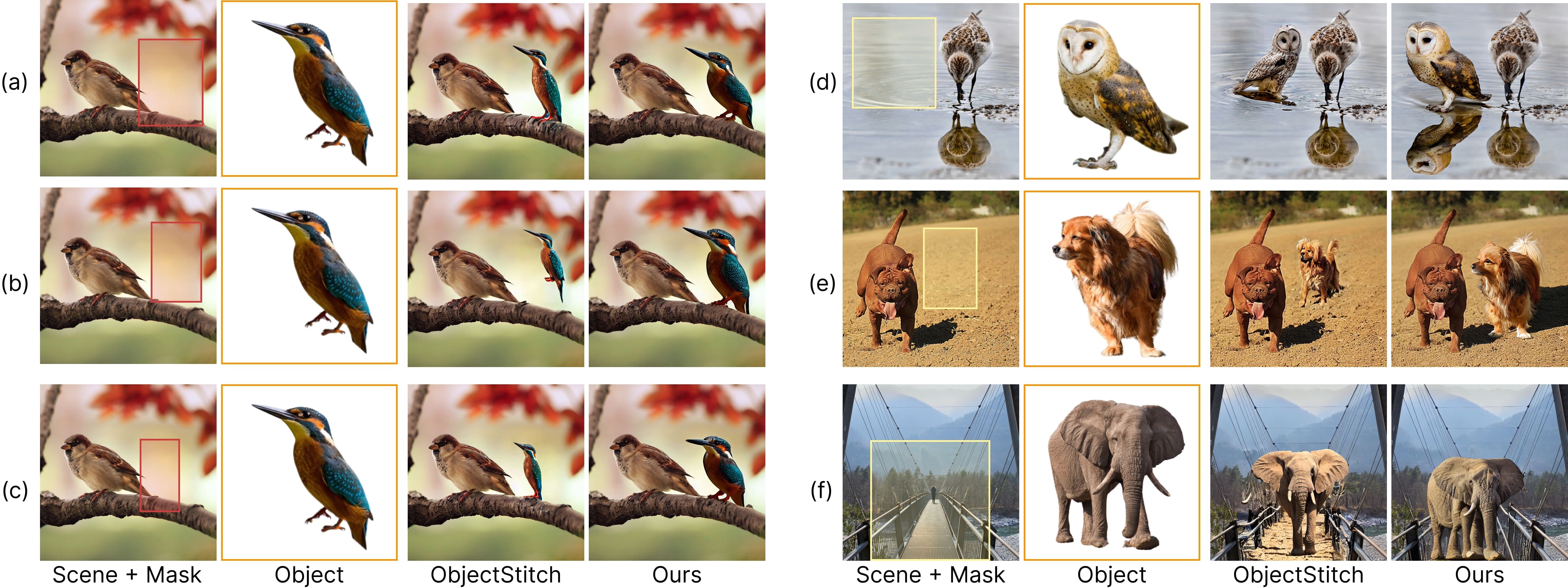}
    \caption{Visual comparison of our model against ObjectStitch \cite{song2022objectstitch}, exemplifying the main benefits of introducing \textit{unconstrained image compositing} as a novel task. (a-c) show different perturbations of the same bbox (red) leading to unnatural results in ObjectStitch; (d-e) Our model can produce more realistic compositions by allowing shadows/reflections beyond the bbox; (f) due to masking the bbox region on the background during training, prior models can create visible changes in the background surrounding the object while our model ensures better preservation.}
    \label{fig:suppmotivation}
\end{figure*}

\section{Unconstrained Image Compositing}
\label{sec:suppmotivation}

As explained in Section 1, prior generative compositing models tend to restrict the generation to a bounding box provided as part of the input. Hence, the definition of the provided mask becomes a pivotal part of the compositing process. If the region defined in the mask is not completely accurate, providing a bounding box that is slightly misplaced or needs rescaling, the quality of the composition can be highly compromised. In this paper, this limitation is addressed by the introduction of \textit{unconstrained image compositing} as a novel task. By allowing generation outside the bbox, the need for crafting accurate masks disappears, alleviating the user work and accelerating the compositing work flow. Fig \ref{fig:suppmotivation} shows visual examples of the main benefits of using an unconstrained approach: 
(a-c) an inaccurate bbox can lead to unnatural composite results (object floating from the surface, unnatural object scale, etc) when using prior mask-based image compositing models,
(d-e) enabling object effects exceeding the bbox boundaries becomes possible with our model leading to more natural results,
(f) by providing the entire background image as part of the input (instead of a masked background), our model achieves a more faithful background preservation around the object.

\section{Data Generation}
\label{sec:suppdatagen}

We introduce in Section 3.1 a \textit{fully automated} data generation pipeline that obtains synthetic background images by leveraging image inpainting techniques, while carefully considering shadows and reflections. 

\begin{figure}[t]
    \centering
    \includegraphics[width=\linewidth]{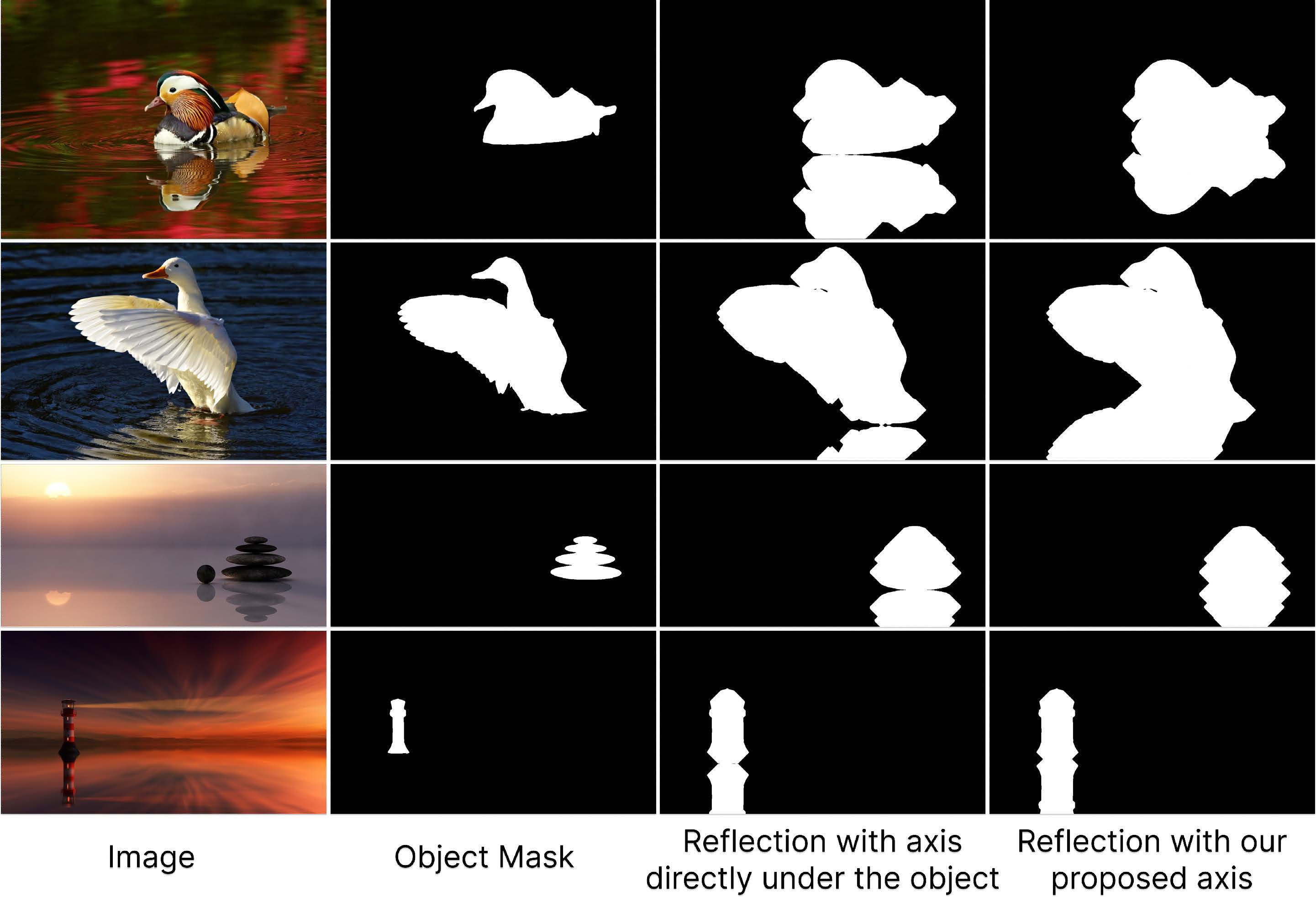}
    \caption{Visualization for the reflection mask in the data generation pipeline. After obtaining each object mask (2nd column), the last two columns show the inpainting mask if the reflection is obtained by flipping the mask right underneath the object (3rd column) or using the axis computed in Eq \ref{eq:refl} (4th column).}
    \label{fig:suppreflection}
\end{figure}

\textbf{Reflection Mask} While an instance shadow detector is used for detecting shadows and estimating their masks, the reflection masks are obtained by using certain heuristics. Only when there is a water body involved in the image, the object mask is vertically flipped to obtain the reflection mask.
However, reflections on water are contingent upon various factors including lighting conditions and camera angle, often resulting in reflections that do not align directly beneath the entire object. This discrepancy is particularly noticeable
when the object's widest part is not what is in contact with water or when the object appears at an angle in the photograph. Therefore, instead of considering the lowest point of the object as the axis of symmetry, we compute it in the following way: (i) We first compute the image row where the object has its widest point as $y_w$. (ii) The lowest pixel of the object is found at image row $y_l$. (iii) The row where we place the horizontal axis is computed as: 

\begin{equation}
    y_{axis} = y_w - \nu(y_l - y_w),
    \label{eq:refl}
\end{equation}

with $\nu = \frac{1}{4}$. Fig \ref{fig:suppreflection} compares the reflection mask obtained using our proposed axis (4th column) to straightforward flipping the object mask right underneath the object (3rd column).

\begin{figure}[t]
    \centering
    \includegraphics[width=\linewidth]{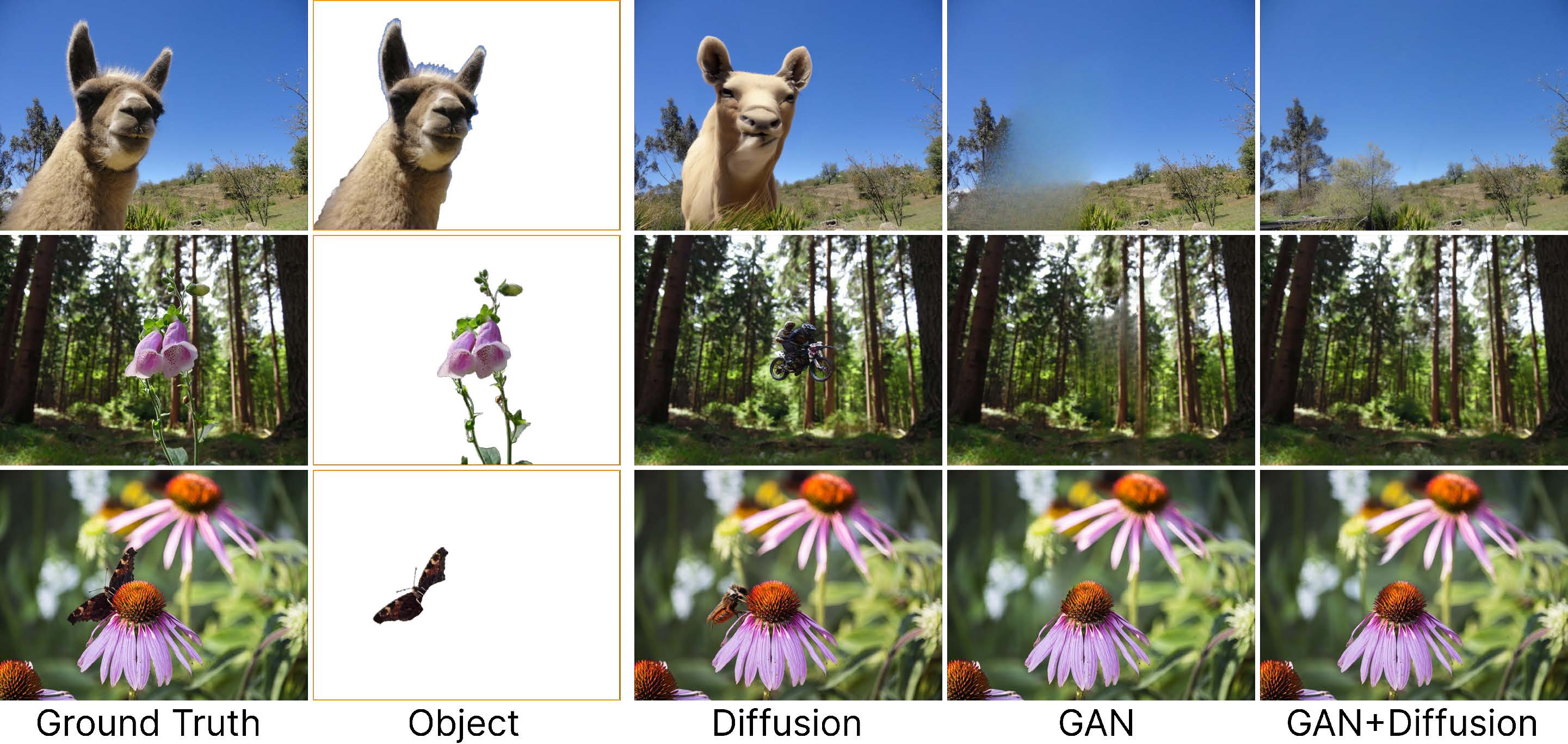}
    \caption{Visualization for the inpainting stage of the data generation pipeline. Inpainting an object (2nd column) on a ground truth image (1st column) via a diffusion-based model \cite{rombach2022ldm} (3rd column) can lead to the synthesis of new objects. GAN-based models \cite{zheng2022cmgan} (4th column) preserve the content but can lead to noticeable artifacts. Setting the parameters of a diffusion-based model to preserve the image content and applying it on the output of a GAN-based model (5th column) achieves a clean background image with no new objects and natural results.}
    \label{fig:suppdataset}
\end{figure}

\textbf{Background inpainting} When the complete inpainting mask is defined, both a GAN-based inpainting model and a diffusion-based one are combined for obtaining the background image with optimal results. By first using a GAN-based inpainting model we ensure obtaining a clean background image where no new objects are introduced in the inpainted region. This image is then refined by a diffusion-based inpainting model for improving the texture and overall generation quality in the image, ensuring a polished and natural-looking background image is synthesized. 
As an example, we demonstrate in Fig \ref{fig:suppdataset} how directly using a diffusion-based model (Stable Diffusion XL Inpainting \cite{rombach2022ldm}) for synthesizing a background image instead of combining both models is prone to filling the inpainted region with new objects instead of providing a clean empty background, motivating the use of our proposed pipeline. After applying the GAN-based inpainting (CM-GAN \cite{zheng2022cmgan}) and obtaining an image with no new object in the inpainted region, we tune the diffusion model settings to preserve the content in the area (noise strength = 0.3), obtaining a refined faithful version of the clean background.

\section{Experiments}

\subsection{Effect of Each Training Stage}
\label{sec:suppmethod}

Our training is performed in stages (Section 3.3) to achieve optimal results. While our model can converge with single-stage training, stage-wise training provides superior composite images and improved training stability. Each stage is detailed for reproducibility and key aspects are analyzed in this section.

\begin{figure}[t]
    \centering
    \includegraphics[width=\linewidth]{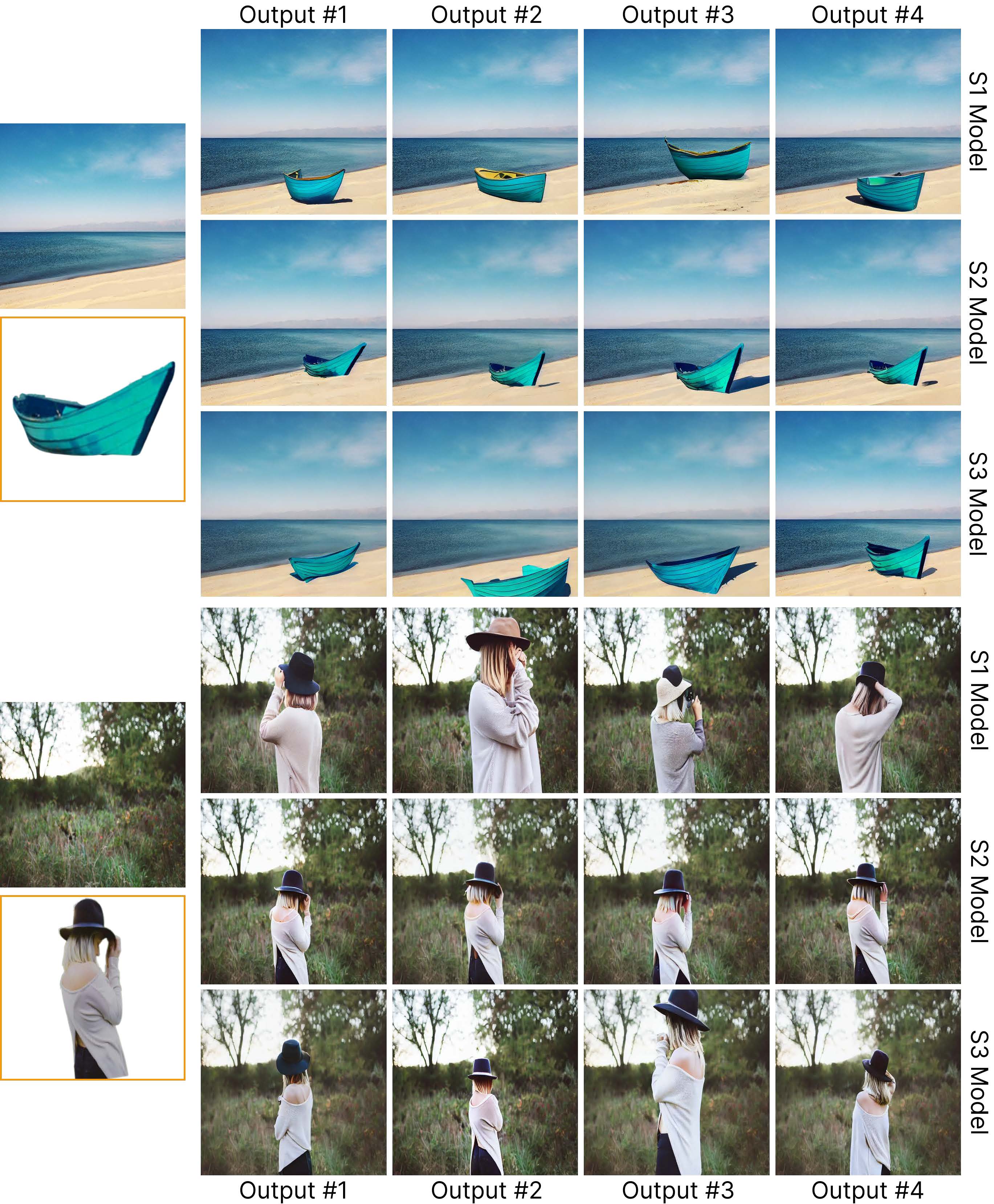}
    \caption{Visualization for the model merging stage. Several diverse outputs are displayed for different ablated models (\textit{S1}, \textit{S2}, \textit{S3}) showcasing the effect of merging the first two models into the third one.}
    \label{fig:suppmerged}
\end{figure}

\textbf{Model Merging} The first two stages consist of training separate models (\textit{S1} and \textit{S2}) and merging them for obtaining a new model (\textit{S3}) that combines the benefits from each of them. As portrayed in Fig \ref{fig:suppmerged}, \textit{S1} model produces composite results with great diversity in pose and location but fails to preserve the key identity features of the object. \textit{S2}, in contrast, is able to reproduce all the key identity features but lacks on diversity. By merging both models into a new \textit{S3} model, we obtain a trade-off between the identity preservation of \textit{S2} and the diversity of \textit{S1}. Note that our final model aims to optimize the trade-off between diversity and identity preservation, but merging weights can be adjusted based on priority.

\begin{figure}[t]
    \centering
    \includegraphics[width=\linewidth]{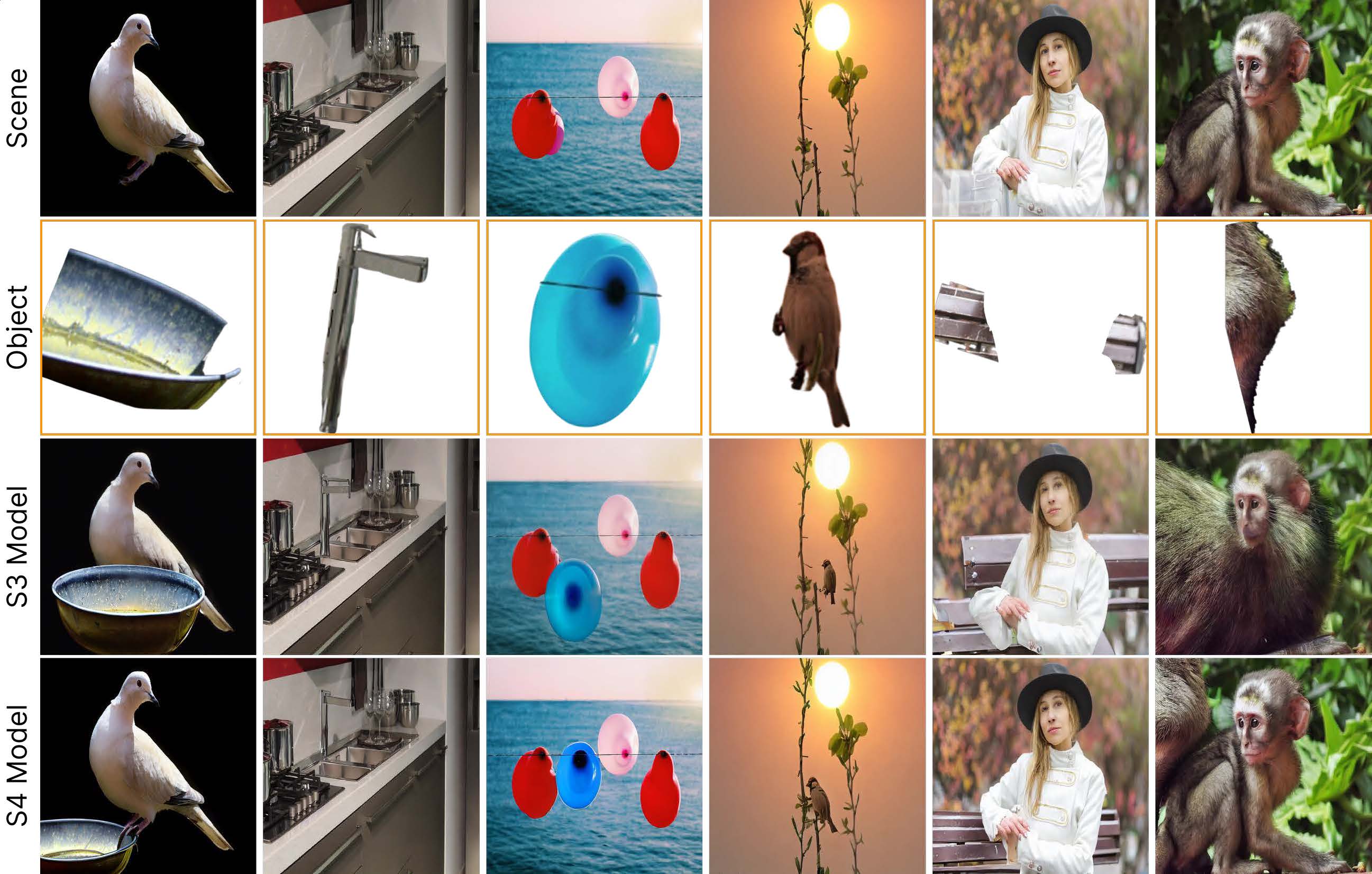}
    \caption{Visualization for the training stage after adding the possibility of providing a rough bbox. During training of \textit{S4}, a rough bounding box is provided with a certain probability while \textit{S3} is only provided with empty masks during training. At inference time, even when providing an empty mask, our \textit{S4} model is able to place the objects in more natural location and scales than \textit{S3} model. }
    \label{fig:supproughbbox}
\end{figure}

\textbf{Optional Input Mask} The next training stage (resulting in \textit{S4} model) consists of providing an input mask containing a rough bounding box with $50\%$ probability, alternating with an empty mask the remaining $50\%$ of the time. This stage allows for the user to provide a bounding box that roughly guides the compositing process without constraining it. Moreover, as observed in Fig \ref{fig:supproughbbox}, even when no mask is provided, it contributes to improving the semantical location and scales of composited objects.

\begin{figure}[t]
    \centering
    \includegraphics[width=\linewidth]{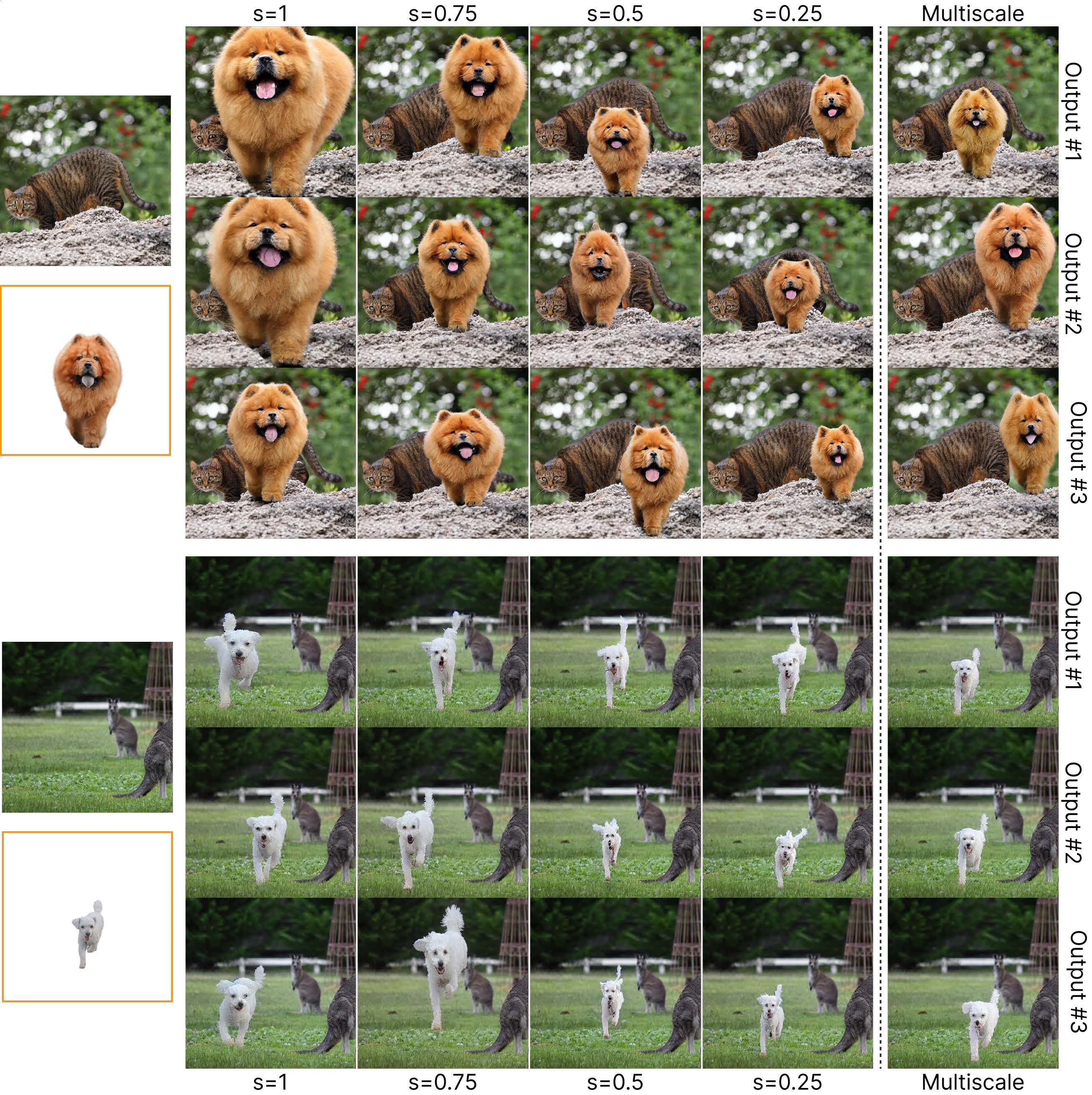}
    \caption{Visualization for the effect of scale on $\mathcal{E}$. In two separate examples, we demonstrate several outputs produced by our model with a fixed scale $s \in \{1, 0.75, 0.5, 0.25\}$ and after introducing a multiscale approach (\textit{S5} model). Our model with multiscale approach is able to obtain a more diverse and accurate outcome.}
    \label{fig:suppmultiscale}
\end{figure}

\textbf{Multiscale Approach} \textit{S5} model is obtained by adding a multiscale approach (Section 3.2) into the training. Fig \ref{fig:suppmultiscale} shows how the encoder is sensitive to the object's scale, leading to composite results where there is a bias towards a certain object scale. By combining the encoding of the object at several scales, our model is able to overcome this bias and produce more diverse results with realistic scales. 

We provide in Tab \ref{tab:multiscale} an additional experiment supporting the design of our multiscale approach. In this experiment, we compare two ways of combining the four different embeddings encoding information of the object $\mathcal{O}$ at different scales: via averaging and via concatenation. In both cases, each embedding is obtained by encoding the object $\mathcal{O}_s$ at a scale $s\in\{1, 0.75, 0.5, 0.25\}$ using the encoder $\mathcal{E}$, as explained in the paper. As we can see in the table, averaging the different embeddings obtains significantly better results in terms of CLIP-score and DreamSim, and comparable DINO-score. Therefore, it is the combination technique we chose for our final model.

\begin{table}[t!]

\centering
\begin{adjustbox}{}
\begin{tabular}{lccc}
\toprule
\textbf{Method}                           &  \textbf{CLIP-score$\uparrow$} & \textbf{DINO-score$\uparrow$} & \textbf{DreamSim$\downarrow$} \\ 
\cmidrule{1-4}
Concatenation                  &    51.623  &  \textbf{73.851}   &  0.803              \\ 
\cmidrule{1-4}
Average   &       \textbf{69.597}  &  71.527  & \textbf{0.420}                                                                                                        \\ 
\bottomrule
\end{tabular}
\end{adjustbox}

\caption{Quantitative evaluation of two versions of our model. In the first, the embeddings from different scales are combined through concatenation and in the second they are combined by computing their average. All metrics are computed on DreamBooth set and no bounding box is provided as input for any of the experiments.}
\label{tab:multiscale}
\end{table}

\begin{figure}[t]
    \centering
    \includegraphics[width=\linewidth]{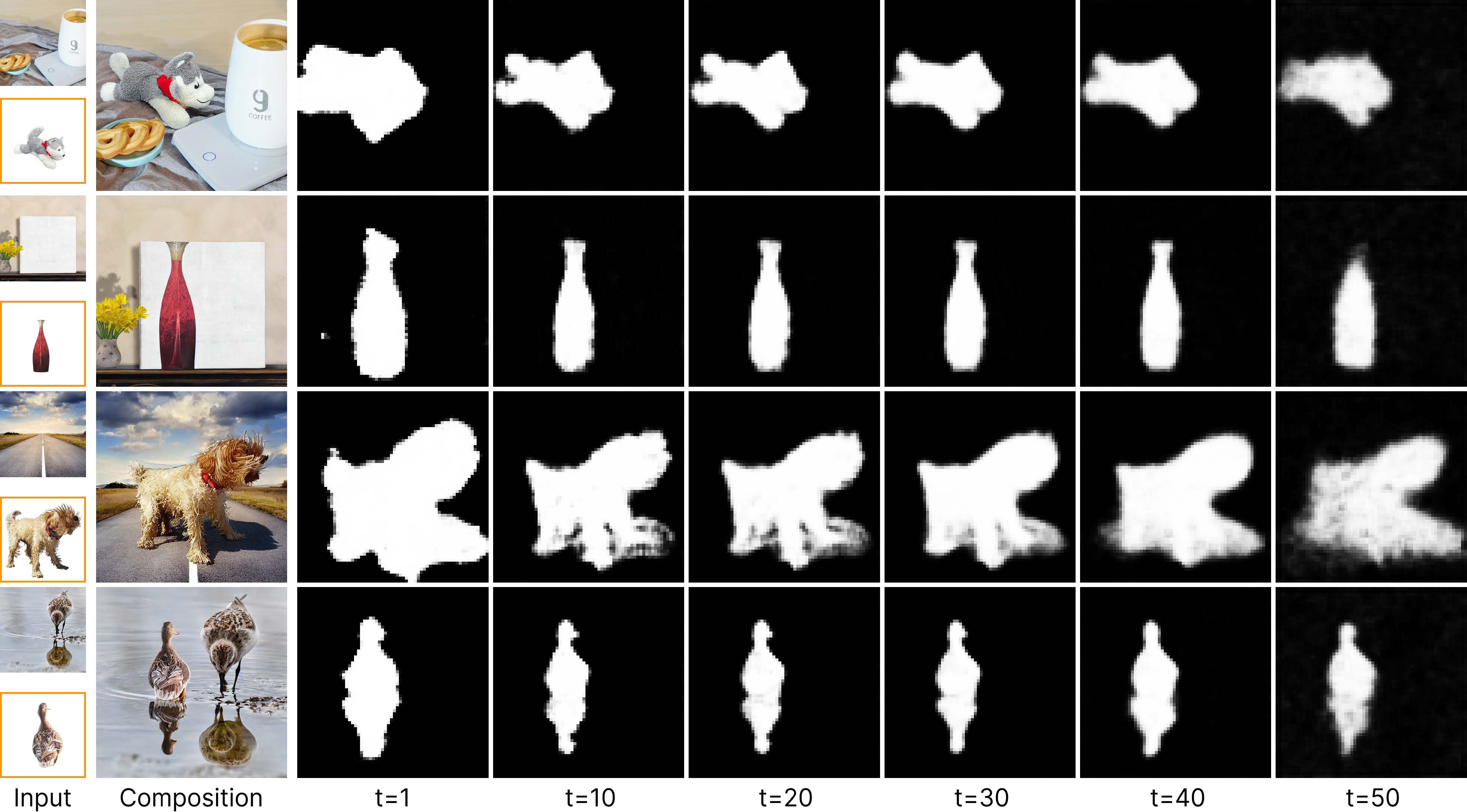}
    \caption{Visualization of Composite Image and Mask produced by our model. For each generated image (obtained in 50 timesteps), the mask obtained at different timesteps is displayed.}
    \label{fig:suppmasks}
\end{figure}

\textbf{Predicted Mask} The last training stage consists of adapting the model for producing an extra output by only training the additional weights while the rest of the network is freezed. Thus, our final model not only generates a composition of the object into the background, but it also provides a mask locating all of the changed pixels. This mask includes the detection of the object in the new image, as well as any shadow, reflection or object effect. At inference time, generating a new image requires 50 timesteps in the reverse diffusion process, while a usable mask can be obtained in only 10 timesteps and becomes optimal at around 10-30 timesteps (Fig \ref{fig:suppmasks}).

\subsection{Mask-free Composite Image Generation}

\begin{figure*}[t]
    \centering
    \includegraphics[width=\linewidth]{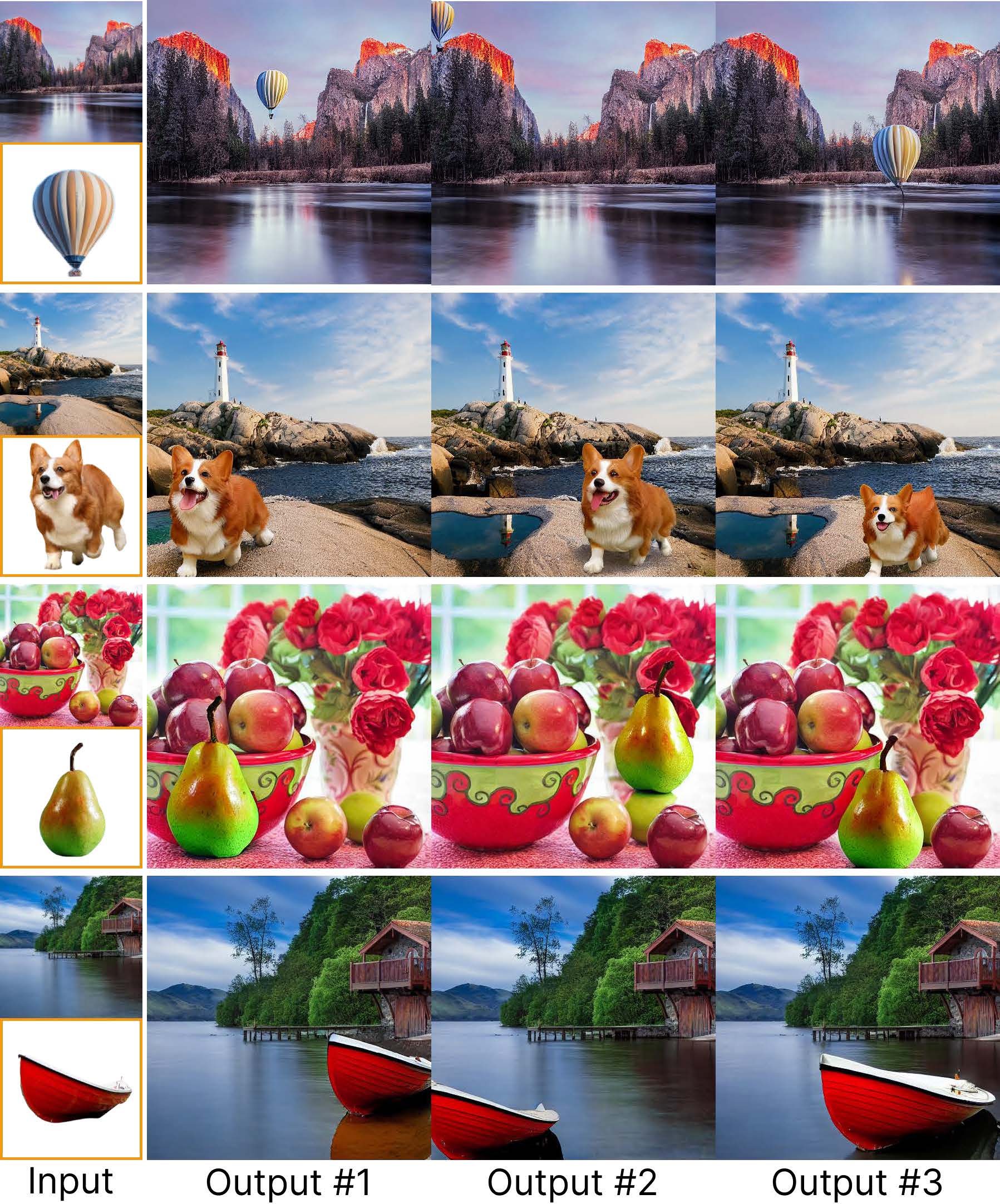}
    \caption{Additional visualization of diverse compositions obtained by employing our model with an empty input mask.}
    \label{fig:suppdiversity}
\end{figure*}

We showcase in Fig \ref{fig:suppdiversity} additional examples
of our model generated with an empty mask. In this case,
the model automatically places the object in a natural location and scale. This feature can also be used as a tool for obtaining ideas of diverse ways to combine the object and the background image.

\subsection{Effect of Bounding Box Scale}

\begin{figure*}[t]
    \centering
    \includegraphics[width=\linewidth]{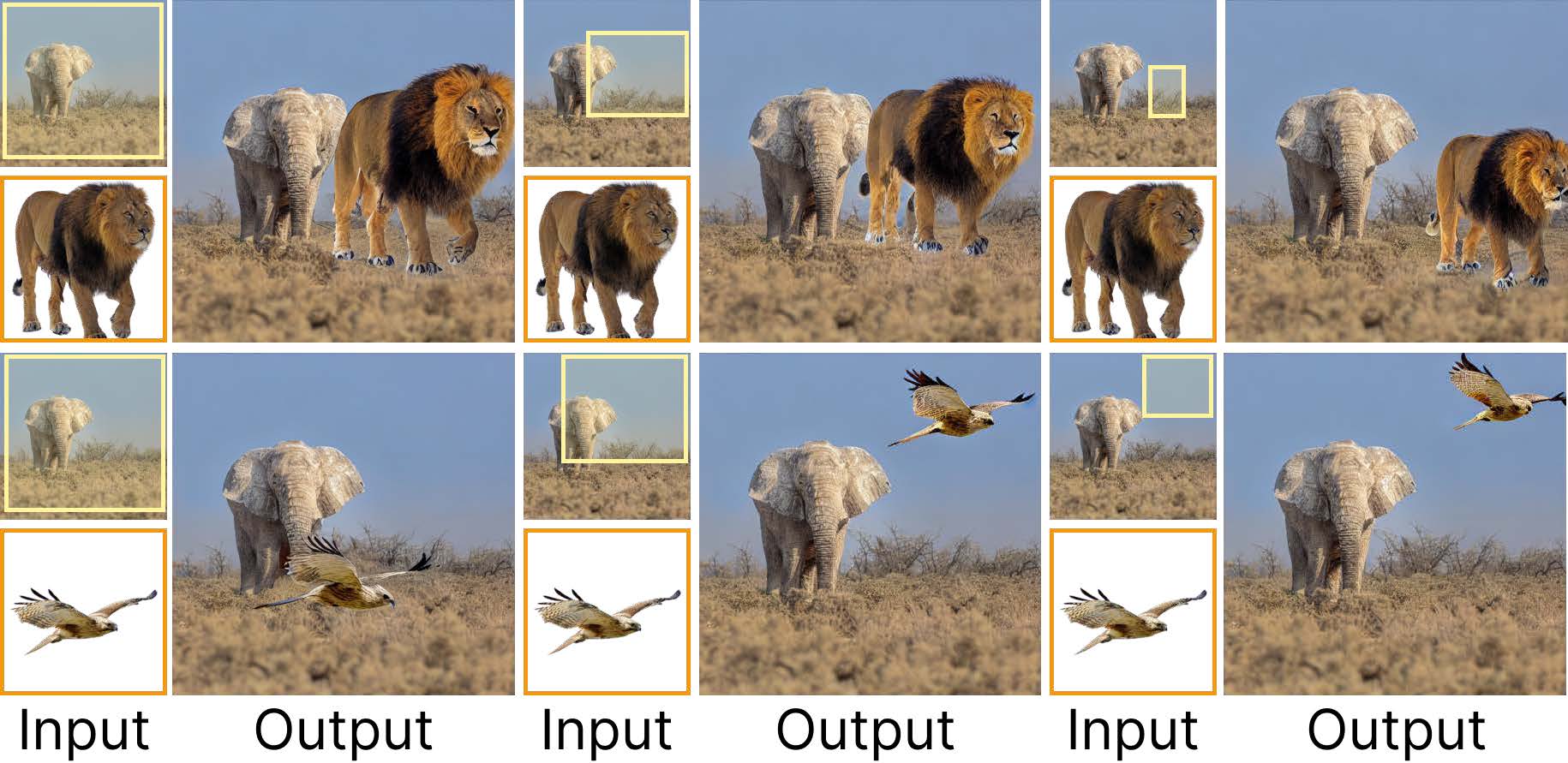}
    \caption{Visualization of compositing a large (top) and a small (bottom) object into the same scene using different sized bounding boxes. The model tends to produce relatively natural object scales even if the bounding box's scale is not natural, using it as a rough guidance. When the bounding box has a natural scale, the object follows it more tightly.}
    \label{fig:suppbbox}
\end{figure*}

Our model is designed to generate realistic composite images whether a bounding box is provided as input or not. As depicted in Fig \ref{fig:suppbbox}, even when an unnaturally large or small bounding box is provided, the model produces a relatively natural object scale, using the provided mask as a rough guidance. If the scale of the provided bounding box is already natural, the object more closely adheres to it.

\subsection{Additional Examples In Challenging Scenarios}

From diverse training data, our model learns to harmonize color, lighting, and geometry (view and pose) and generate object effects that fit the background scene.

\begin{figure*}[t]
    \centering
    \includegraphics[width=\linewidth]{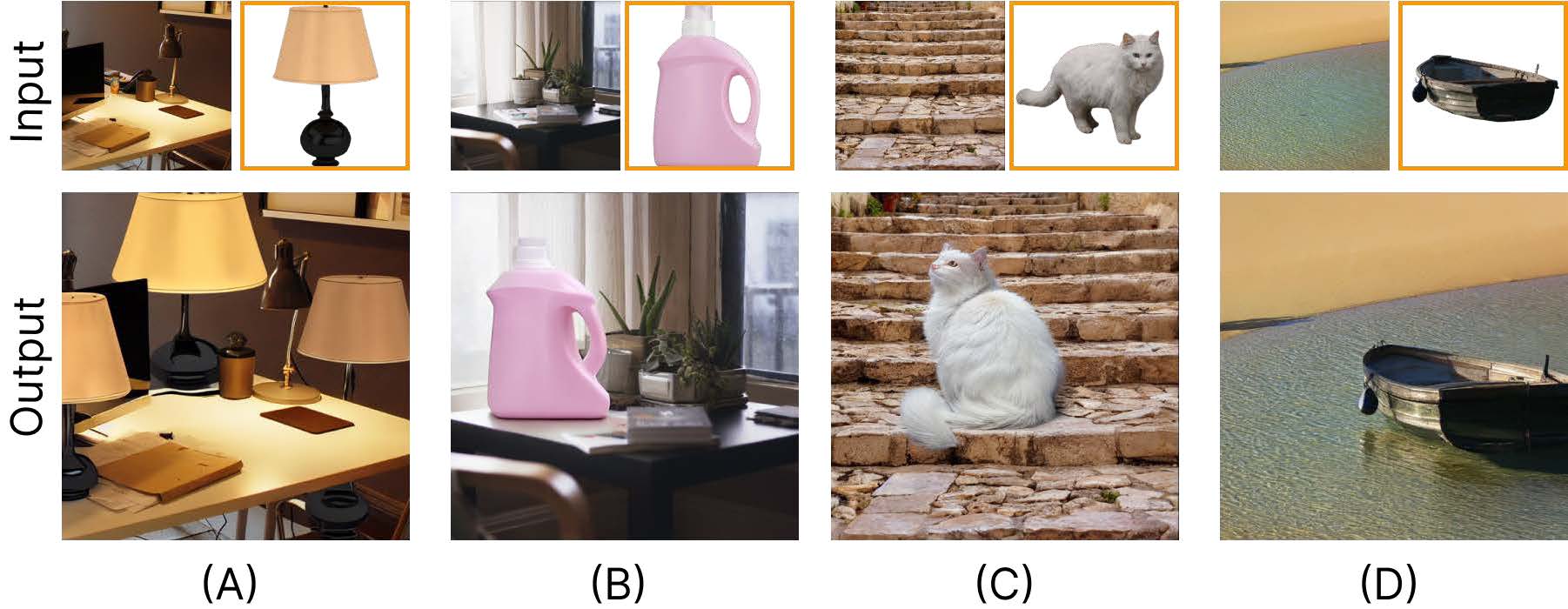}
    \caption{Visualization of compositions obtained in different complex scenarios. (A) object (lamp) is composed into three different locations in a scene with complex lighting, resulting in three different relightings; (B) the object is composed with a material-aware relighting and harmonization; (C) object is reposed by the model for obtaining a more natural scene; (D) shadows and reflections of the object are simultaneously generated by the model to provide a natural composition.}
    \label{fig:suppcomplex}
\end{figure*}

Despite not explicitly estimating lighting, the model uses the diverse training data to infer scene lighting, allowing for appropriate object relighting. In Fig \ref{fig:suppcomplex} (A), we show the same object placed in three different locations within the same scene with complex lighting, resulting in three different relightings. Fig \ref{fig:suppcomplex} (B) demonstrates material harmonization, \ref{fig:suppcomplex} (C) shows reposing and \ref{fig:suppcomplex} (D) illustrates simultaneous shadow and reflection generation. 


\section{Comparison to Existing Methods}
\label{sec:suppbaselines}

\subsection{User Studies}
\label{sec:suppuserstudies}

\begin{figure}[t]
    \centering
    \includegraphics[width=\linewidth]{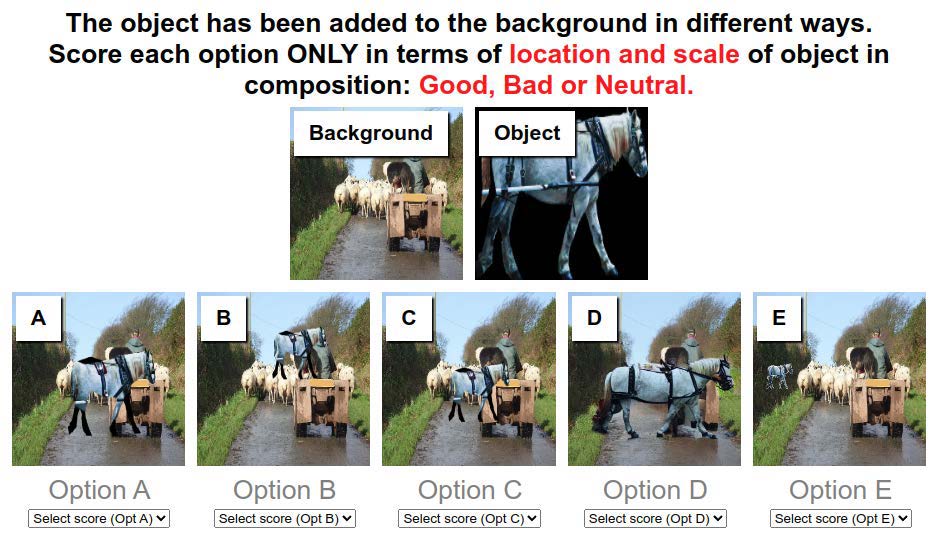}
    \caption{Screenshot of user study presented to users for evaluating location and scale against object placement prediction baselines.}
    \label{fig:suppuserstudypos}
\end{figure}

\begin{figure}[t]
    \centering
    \includegraphics[width=0.8\linewidth]{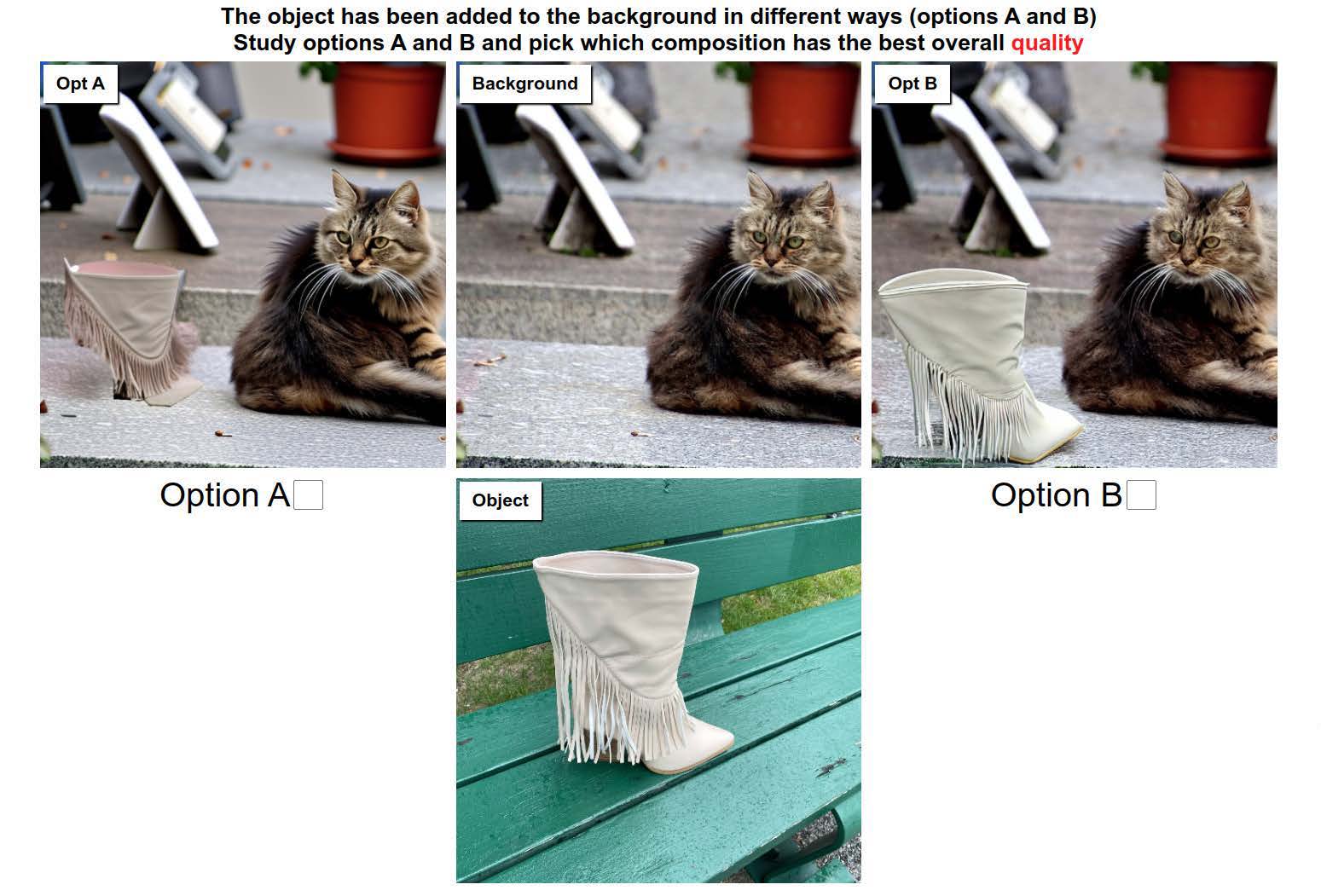}
    \caption{Screenshot of user study presented to users for evaluating output quality against generative image compositing baselines.}
    \label{fig:suppuserstudyqual}
\end{figure}

\begin{figure}[t]
    \centering
    \includegraphics[width=0.8\linewidth]{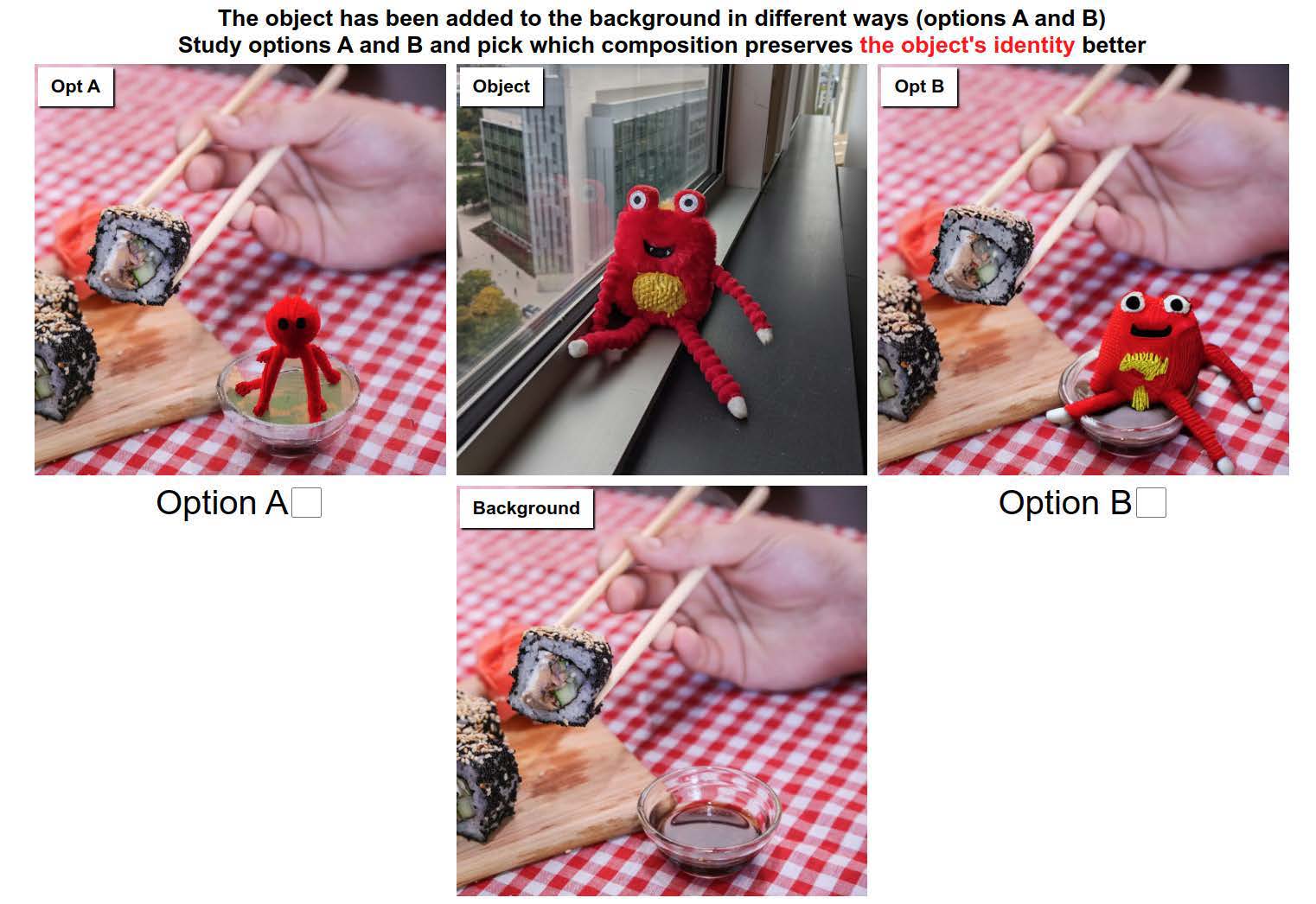}
    \caption{Screenshot of user study presented to users for evaluating identity preservation against generative image compositing baselines.}
    \label{fig:suppuserstudyid}
\end{figure}

\begin{figure}[t]
    \centering
    \includegraphics[width=0.8\linewidth]{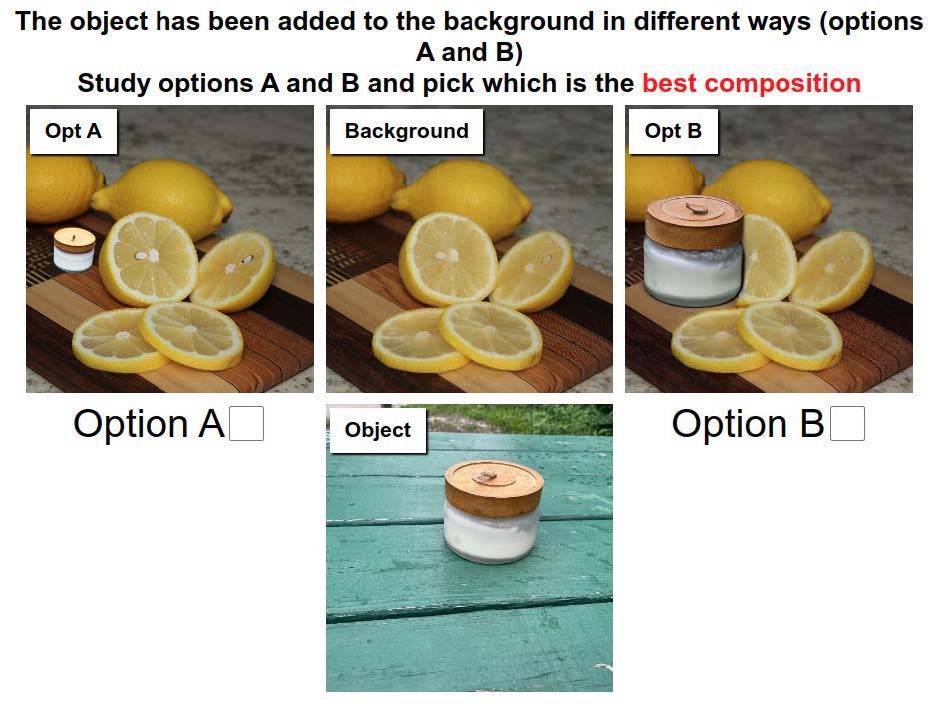}
    \caption{Screenshot of user study presented to users for evaluating automated object compositing via our model vs a two-step pipeline.}
    \label{fig:suppuserstudy2stage}
\end{figure}

As mentioned in Section 4.1, we conducted several user studies to visually evaluate our model against different sets of baselines. In each of them, at least 5 users were considered, and preferences were measured via majority consensus voting. Those experiments consist of:

\begin{itemize}
    \item \textbf{Object Placement Prediction.} We compare our model to TopNet \cite{zhu2023topnet}, GracoNet \cite{zhou2022graconet}, PlaceNet \cite{zhang2020placenet} and TERSE \cite{tripathi2019terse} on a subset of 150 images from the OPA \cite{liu2021opa} test set. For each input, we display the output of each model side by side in a randomized order and ask the user to rate each image individually as Good, Bad or Neutral. We specifically ask the rating to be based on \textbf{ONLY} location and scale. We show an example of this user study in Fig \ref{fig:suppuserstudypos}.
    \item \textbf{Compositing Quality.} We compare to diffusion-based image compositing models Paint by Example \cite{yang2023paintbyexample}, ObjectStitch \cite{song2022objectstitch}, TF-ICON \cite{lu2023tficon}, AnyDoor \cite{chen2023anydoor}, ControlCom \cite{zhang2023controlcom} in terms of \textit{compositing quality}. For each of the 113 (background image, object, ground-truth bbox mask) triplets that constitute our DreamBooth test set, we show side by side our output and one of the baselines, in randomized order. The user is asked to choose the best composition in each case, in terms of \textit{quality} (Fig \ref{fig:suppuserstudyqual}).
    \item \textbf{Identity Preservation.} We perform a human study with the same settings as the previous one. This time, however, users are asked to pick their preference in terms of \textit{identity preservation} (Fig \ref{fig:suppuserstudyid}).
    \item \textbf{Automatic Image Compositing Pipeline: End-to-end vs Two Stages.} We ask users to express their preference between (i) using our end-to-end unconstrained pipeline, in which an empty mask is provided and the object is generated in a location and scale simultaneously predicted by the model, and (ii) two-stage compositing pipeline by employing a diffusion-based object compositing model (ObjectStitch \cite{song2022objectstitch}, Paint by Example \cite{yang2023paintbyexample}, TF-ICON \cite{lu2023tficon}, AnyDoor \cite{chen2023anydoor}, ControlCom \cite{zhang2023controlcom}) using as input a bounding box predicted by TopNet \cite{zhu2023topnet}, an object placement prediction model. For each pair of background and object images in the DreamBooth test set, the output of each method is computed and shown side by side to the user in a randomized order. The user is then asked to pick their preferred composition out of the two of them (Fig \ref{fig:suppuserstudy2stage}). 
\end{itemize}

\subsection{Comparison to Object Placement Prediction Models}
\label{sec:suppexamplespos}

\begin{figure*}[t]
    \centering
    \includegraphics[width=\textwidth]{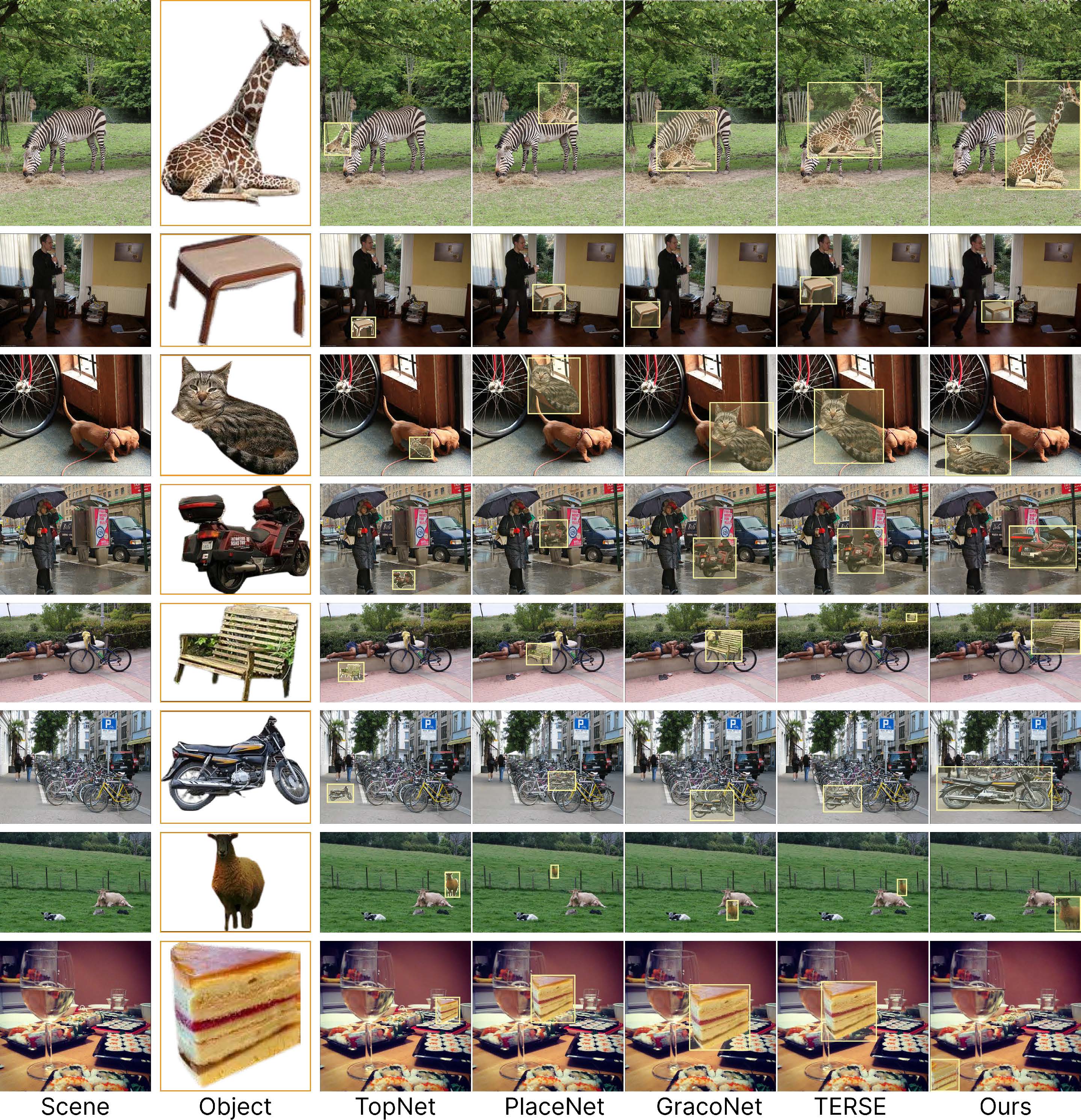}
    \caption{Visual comparison between our model and Object Placement Prediction Models. Note that for all baselines (TopNet \cite{zhu2023topnet}, PlaceNet \cite{zhang2020placenet}, GracoNet \cite{zhou2022graconet} and TERSE \cite{tripathi2019terse}), the object is copy-pasted into the predicted location for easier visual comparison.}
    \label{fig:suppbaselinespos}
\end{figure*}

We provide additional visualization examples comparing our model to TopNet \cite{zhu2023topnet}, PlaceNet \cite{zhang2020placenet}, GracoNet \cite{zhou2022graconet} and TERSE \cite{tripathi2019terse} in Fig \ref{fig:suppbaselinespos}.

\subsection{Comparison to Generative Image Compositing Models}
\label{sec:suppexamplesgen}

\begin{figure*}[t]
    \centering
    \includegraphics[width=\textwidth]{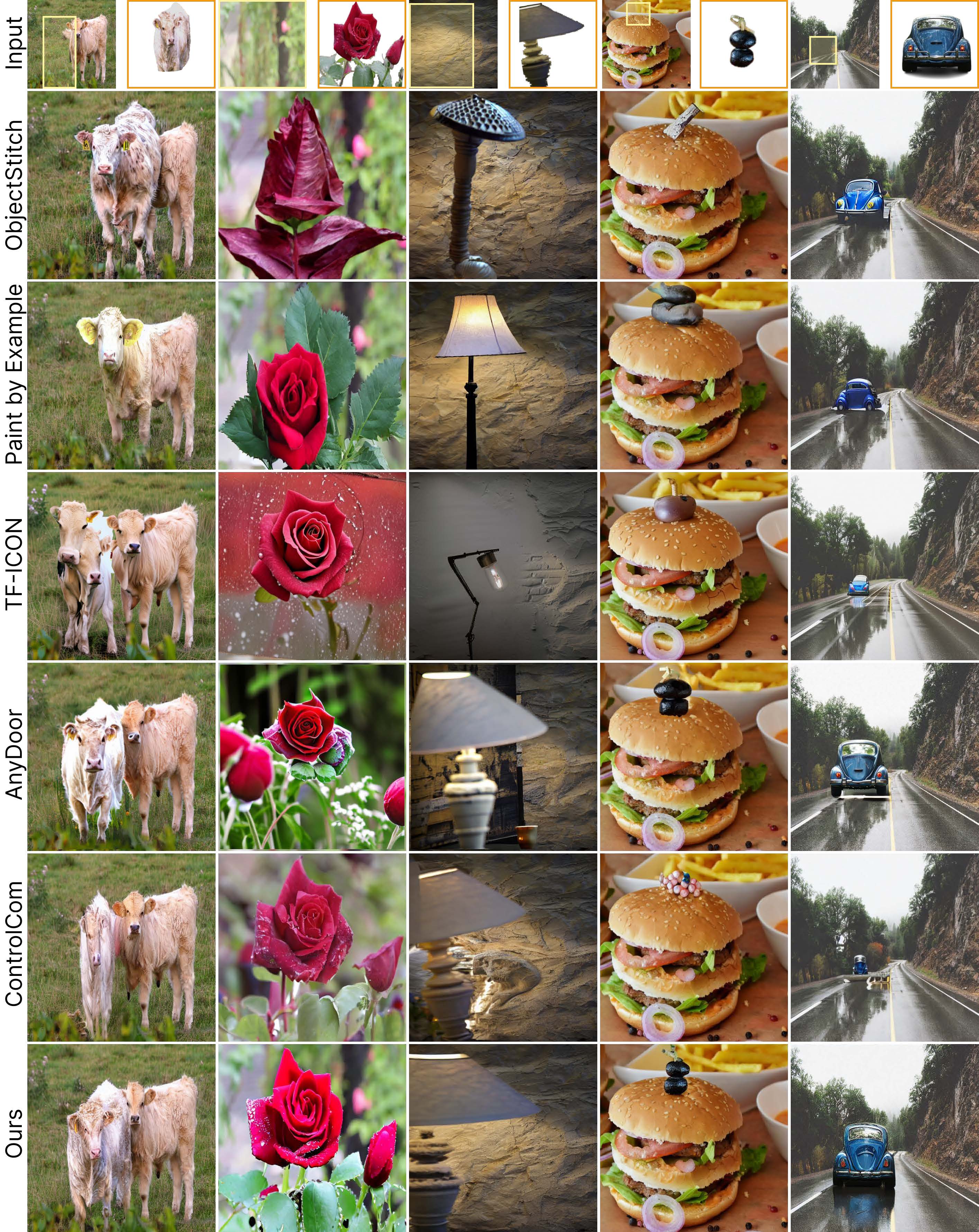}
    \caption{Visual comparison between our model and different Generative Image Compositing models (ObjectStitch \cite{song2022objectstitch}, Paint By Example \cite{yang2023paintbyexample}, TF-ICON \cite{lu2023tficon}, AnyDoor \cite{chen2023anydoor}, ControlCom \cite{zhang2023controlcom}).}
    \label{fig:suppbaselines1}
\end{figure*}

\begin{figure*}[t]
    \centering
    \includegraphics[width=\textwidth]{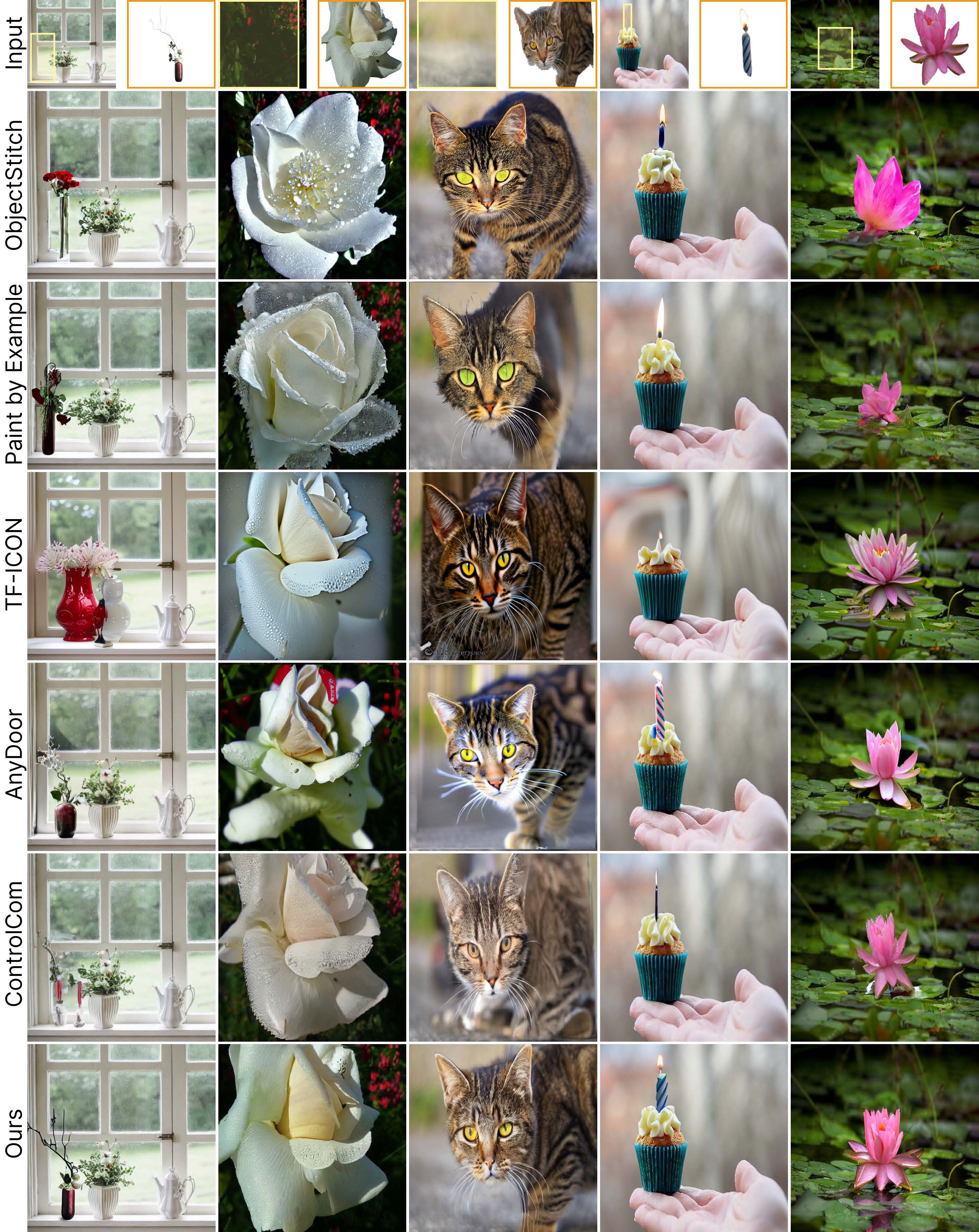}
    \caption{Additional visual comparison between our model and different Generative Image Compositing models (ObjectStitch \cite{song2022objectstitch}, Paint By Example \cite{yang2023paintbyexample}, TF-ICON \cite{lu2023tficon}, AnyDoor \cite{chen2023anydoor}, ControlCom \cite{zhang2023controlcom}).}
    \label{fig:suppbaselines2}
\end{figure*}

\begin{figure*}[t]
    \centering
    \includegraphics[width=\textwidth]{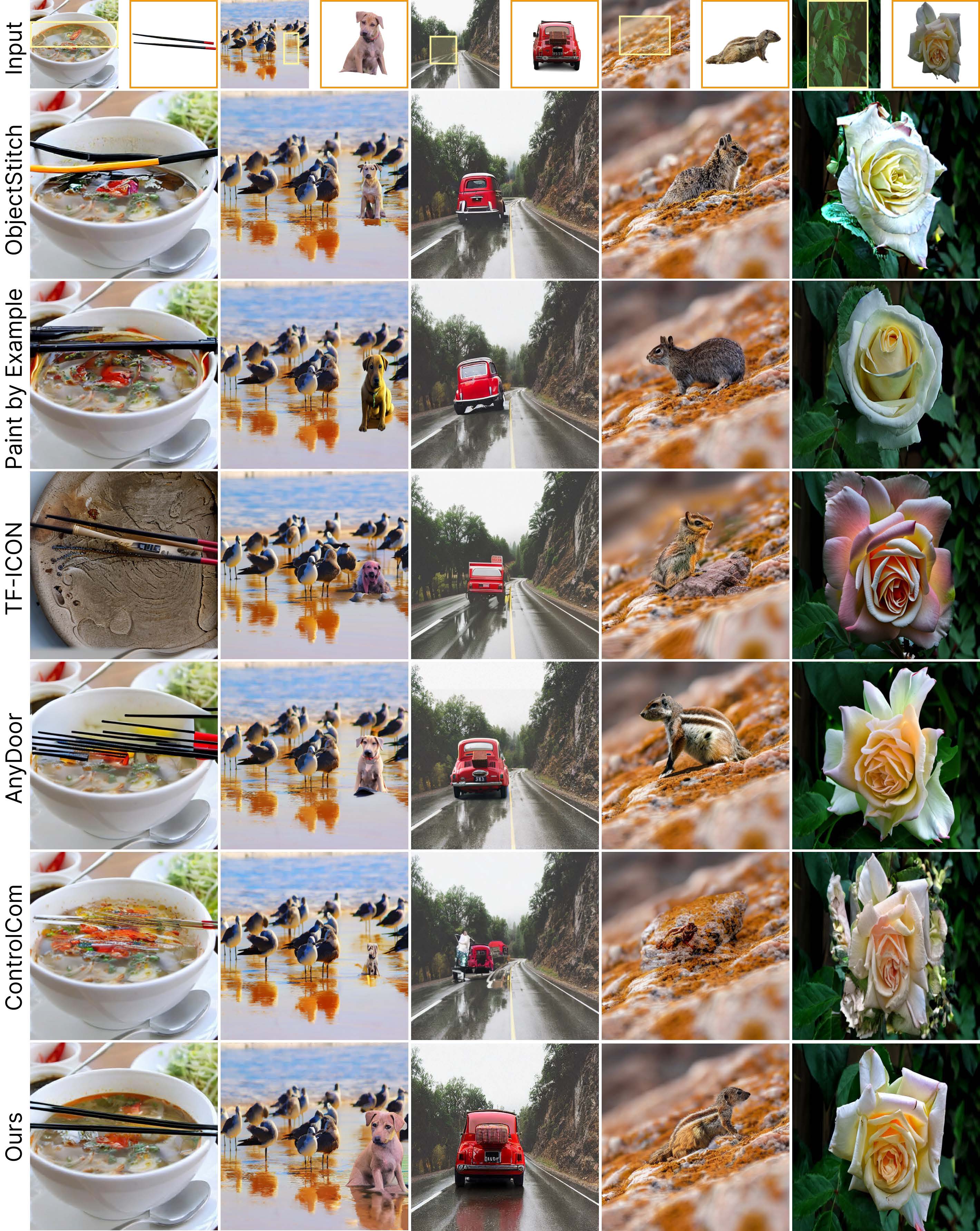}
    \caption{Additional visual comparison between our model and different Generative Image Compositing models (ObjectStitch \cite{song2022objectstitch}, Paint By Example \cite{yang2023paintbyexample}, TF-ICON \cite{lu2023tficon}, AnyDoor \cite{chen2023anydoor}, ControlCom \cite{zhang2023controlcom}).}
    \label{fig:suppbaselines3}
\end{figure*}

\begin{figure*}[t]
    \centering
    \includegraphics[width=\textwidth]{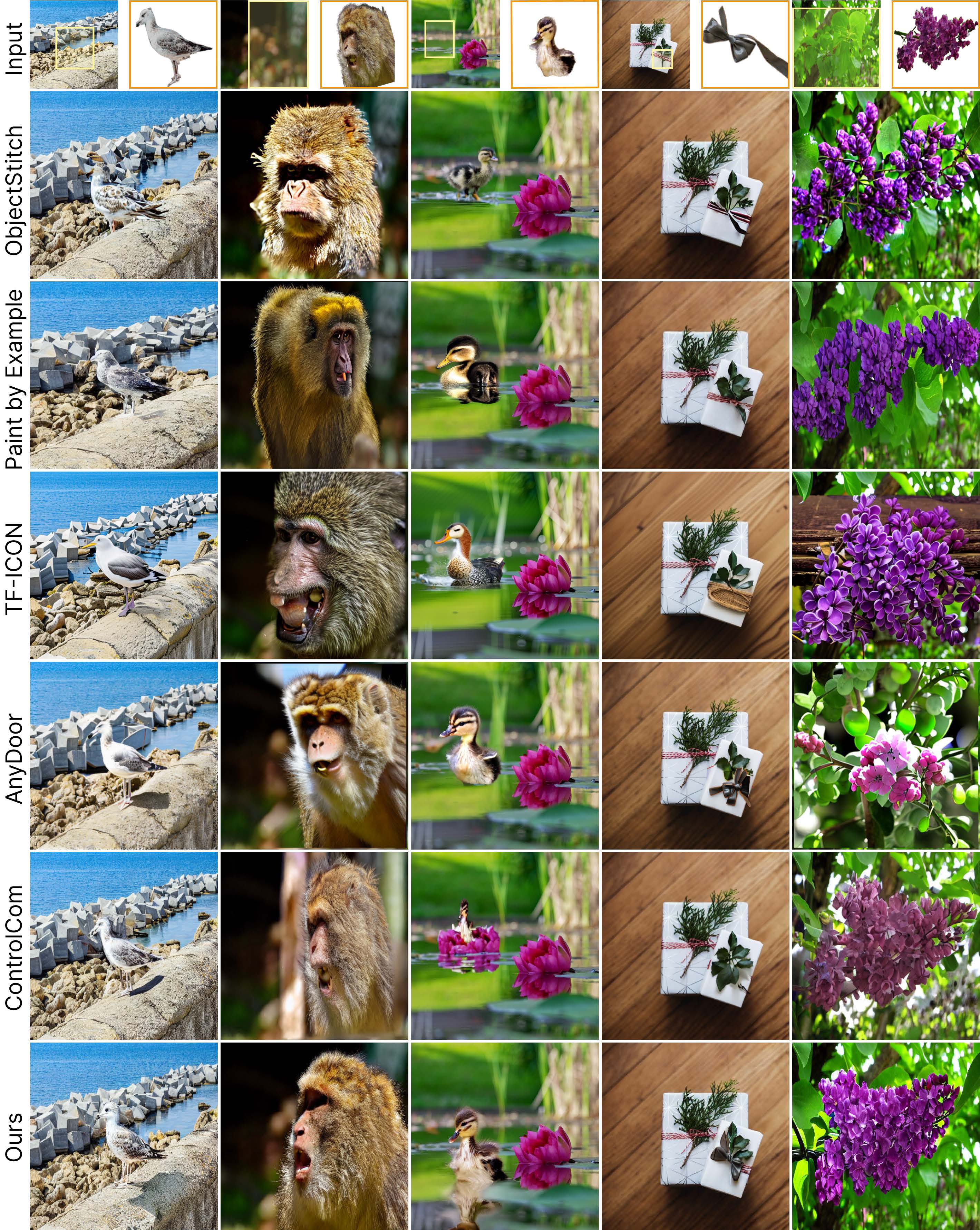}
    \caption{Additional visual comparison between our model and different Generative Image Compositing models (ObjectStitch \cite{song2022objectstitch}, Paint By Example \cite{yang2023paintbyexample}, TF-ICON \cite{lu2023tficon}, AnyDoor \cite{chen2023anydoor}, ControlCom \cite{zhang2023controlcom}).}
    \label{fig:suppbaselines4}
\end{figure*}

\begin{figure*}[t]
    \centering
    \includegraphics[width=\textwidth]{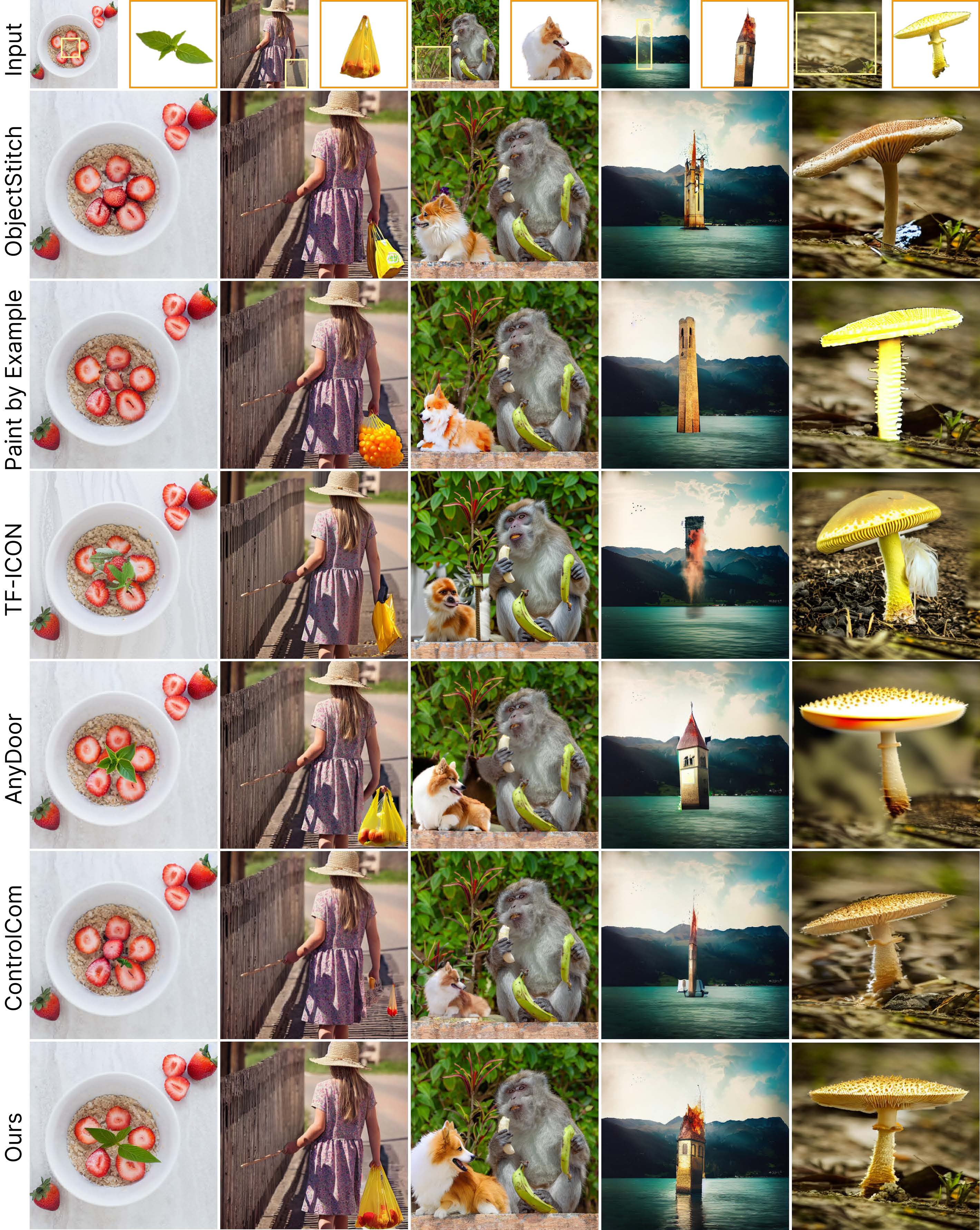}
    \caption{Additional visual comparison between our model and different Generative Image Compositing models (ObjectStitch \cite{song2022objectstitch}, Paint By Example \cite{yang2023paintbyexample}, TF-ICON \cite{lu2023tficon}, AnyDoor \cite{chen2023anydoor}, ControlCom \cite{zhang2023controlcom}).}
    \label{fig:suppbaselines5}
\end{figure*}

We display additional visualization examples comparing our model to prior generative compositing methods (ObjectStitch \cite{song2022objectstitch}, Paint By Example \cite{yang2023paintbyexample}, TF-ICON \cite{lu2023tficon}, AnyDoor \cite{chen2023anydoor}, ControlCom \cite{zhang2023controlcom}) in Figs \ref{fig:suppbaselines1}, \ref{fig:suppbaselines2}, \ref{fig:suppbaselines3}, \ref{fig:suppbaselines4}, \ref{fig:suppbaselines5}.

\subsubsection{Comparison to IMPRINT}

\begin{figure}[t]
    \centering
    \includegraphics[width=\textwidth]{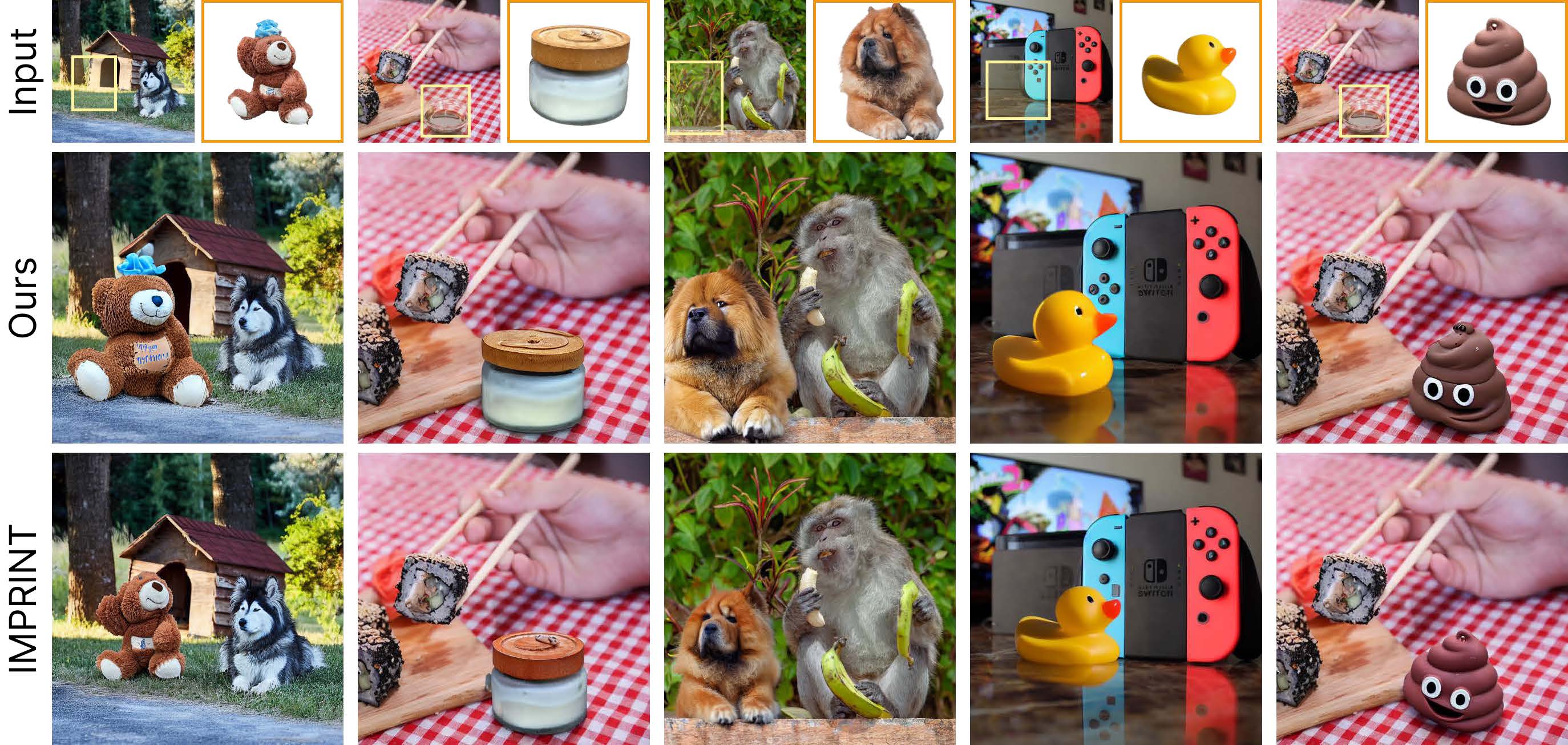}
    \caption{Visual comparison between our model and IMPRINT \cite{song2024imprint}. IMPRINT is a mask-based Generative Image Compositing model released after paper submission, thus not included in prior experiments.}
    \label{fig:suppimprint}
\end{figure}

IMPRINT \cite{song2024imprint} is a generative image compositing model that was not released at the time of paper submission, thus is not included in the initial comparisons. For transparency, we provide a comparison with our model in Tab \ref{tab:suppimprint} and Fig \ref{fig:suppimprint}.

Both IMPRINT and our model utilize multiview and video data in their training, following the methodology described in \cite{song2024imprint}. Tab 4 in \cite{song2024imprint} shows the performance benefits of integrating this data.

Although IMPRINT performs slightly better than our model (Tab \ref{tab:suppimprint}) in some identity and semantics preservation metrics, it should be noted that, unlike IMPRINT (mask-based model), our model focus on \textit{unconstrained} image compositing, thus being able of simultaneously performing object compositing and placement prediction without bounding boxes. Our model trains with full background images instead of masked ones, requiring specific data crafting (Section 3.1). This leads to improved background preservation, more natural object effects, and less reliance on precise bounding boxes than mask-base object compositing models like IMPRINT (See Sec \ref{sec:suppmotivation}). Therefore, while IMPRINT might obtain slightly superior identity preservation, our model offers different advantages with comparable quality and identity preservation.

\begin{table}[t!]
    \centering
    \begin{adjustbox}{width=\textwidth}
    \begin{tabular}{lccccccc}
    \toprule
    \multirow{2}[3]{*}{\textbf{Method}} & \multicolumn{3}{c}{\textbf{DreamBooth}} & \multicolumn{4}{c}{\textbf{Pixabay-Comp}} \\
    \cmidrule(lr){2-4} \cmidrule(lr){5-8}  & \textbf{CLIP-Score$\uparrow$} & \textbf{DINO-Score$\uparrow$} & \textbf{DreamSim$\downarrow$} & \textbf{FID$\downarrow$} & \textbf{CLIP-Score$\uparrow$} & \textbf{DINO-Score$\uparrow$} & \textbf{DreamSim$\downarrow$}\\ 
    \cmidrule{1-8}
    IMPRINT \cite{song2024imprint}                                                                                                           &      \textbf{82.965}                   &  \textbf{88.910}   &           \textbf{0.246}                  &                                                           \textbf{62.026}                                         &           76.874          &  79.465   &             0.435        \\
    \cmidrule{1-8}
    Ours (w/ bbox)                                                                                           &    80.946      &  85.646   &       0.285   &   62.406      &                                                          \textbf{77.129}             &       \textbf{80.896}             &    \textbf{0.395}              \\ 
    
    \bottomrule
    \end{tabular}
    \end{adjustbox}
    \caption{Quantitative comparison of composition quality and identity preservation to IMPRINT \cite{song2024imprint}. FID is only computed on Pixabay-Comp, which has ground truth images. Our model obtains comparable performance with the additional benefits of the \textit{unconstrained generative object compositing} approach.}
    \label{tab:suppimprint}
\end{table}


\subsection{Comparison to Automated Placement and Compositing Pipeline}
\label{sec:suppexamples2step}

\begin{figure}[t]
    \centering
    \includegraphics[width=\linewidth]{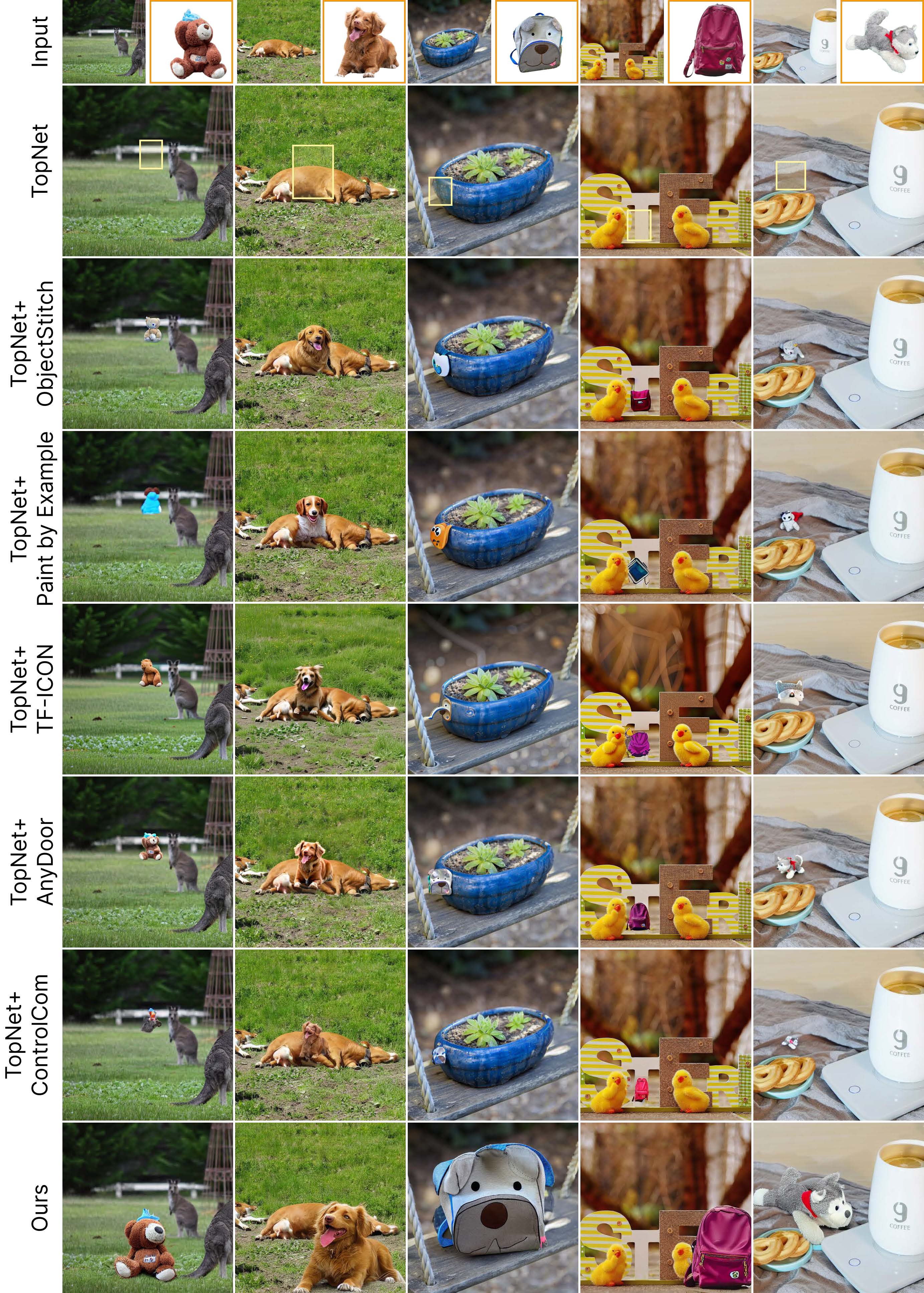}
    \caption{Visualization of mask-free compositing via our model vs a two-step pipeline. For the latter, we use masks predicted by TopNet \cite{zhu2023topnet} (top row) for image compositing via ObjectStitch \cite{song2022objectstitch}, Paint by Example \cite{yang2023paintbyexample}, TF-ICON \cite{lu2023tficon}, AnyDoor \cite{chen2023anydoor} and ControlCom \cite{zhang2023controlcom}.}
    \label{fig:supp2steps}
\end{figure}

We display visualization examples comparing our automated pipeline to a two-step pipeline that combines Object Placement Prediction and Generative Compositing models in Fig \ref{fig:supp2steps}. We use our model in mask-free mode, jointly performing object placement and compositing, and compare to sequentially connecting TopNet \cite{zhu2023topnet} with ObjectStitch \cite{song2022objectstitch}, Paint by Example \cite{yang2023paintbyexample}, TF-ICON \cite{lu2023tficon}, AnyDoor \cite{chen2023anydoor} and ControlCom \cite{zhang2023controlcom}.

\end{document}